\documentclass[letterpaper,twocolumn,english]{IEEEtran}
\usepackage[T1]{fontenc}
\usepackage[latin9]{inputenc}
\usepackage{array}
\usepackage{float}
\usepackage{mathrsfs}
\usepackage{bm}
\usepackage{amsthm}
\usepackage{amsmath}
\usepackage{amssymb}
\usepackage{graphicx}

\makeatletter

\providecommand{\tabularnewline}{\\}
\floatstyle{ruled}
\newfloat{algorithm}{tbp}{loa}
\providecommand{\algorithmname}{Algorithm}
\floatname{algorithm}{\protect\algorithmname}

\theoremstyle{plain}
\newtheorem{thm}{\protect\theoremname}
\theoremstyle{remark}
\newtheorem{rem}[thm]{\protect\remarkname}
\theoremstyle{plain}
\newtheorem{lem}[thm]{\protect\lemmaname}

\@ifundefined{showcaptionsetup}{}{%
 \PassOptionsToPackage{caption=false}{subfig}}
\usepackage{subfig}
\makeatother

\usepackage{babel}
\providecommand{\lemmaname}{Lemma}
\providecommand{\remarkname}{Remark}
\providecommand{\theoremname}{Theorem}

\begin{document}

\title{Simultaneous Codeword Optimization (SimCO) for Dictionary Update
and Learning}

\author{Wei Dai, \emph{Member IEEE}, Tao Xu, \emph{Student Member IEEE},
Wenwu Wang, \emph{Senior Member IEEE}}
\maketitle
\begin{abstract}
We consider the data-driven dictionary learning problem. The goal
is to seek an over-complete dictionary from which every training signal
can be best approximated by a linear combination of only a few codewords.
This task is often achieved by iteratively executing two operations:
sparse coding and dictionary update. In the literature, there are
two benchmark mechanisms to update a dictionary. The first approach,
such as the MOD algorithm, is characterized by searching for the optimal
codewords while fixing the sparse coefficients. In the second approach,
represented by the K-SVD method, one codeword and the related sparse
coefficients are simultaneously updated while all other codewords
and coefficients remain unchanged. We propose a novel framework that
generalizes the aforementioned two methods. The unique feature of
our approach is that one can update an arbitrary set of codewords
and the corresponding sparse coefficients simultaneously: when sparse
coefficients are fixed, the underlying optimization problem is similar
to that in the MOD algorithm; when only one codeword is selected for
update, it can be proved that the proposed algorithm is equivalent
to the K-SVD method; and more importantly, our method allows us to
update all codewords and all sparse coefficients simultaneously, hence
the term \emph{simultaneous codeword optimization (SimCO)}. Under
the proposed framework, we design two algorithms, namely, primitive
and regularized SimCO. We implement these two algorithms based on
a simple gradient descent mechanism. Simulations are provided to demonstrate
the performance of the proposed algorithms, as compared with two baseline
algorithms MOD and K-SVD. Results show that regularized SimCO is particularly
appealing in terms of both learning performance and running speed. 
\end{abstract}
\renewcommand{\thefootnote}{\fnsymbol{footnote}} 
\setcounter{footnote}{-1}

\footnote{This work was supported by the MOD University Defence Research Centre
(UDRC) in Signal Processing.

W. Dai is with the Department of Electrical and Electronic Engineering,
Imperial College London, London SW7 2AZ, United Kingdom (email: wei.dai1@imperial.ac.uk).

T. Xu and W. Wang are with the Department of Electronic Engineering,
University of Surrey, Guildford GU2 7XH, United Kingdom (email: t.xu
and w.wang@surrey.ac.uk).%
}

\renewcommand{\thefootnote}{\arabic{footnote}}
\setcounter{footnote}{0}

\section{\label{sec:intro}Introduction}

Sparse signal representations have recently received extensive research
interests across several communities including signal processing,
information theory, and optimization \cite{Foldiak1990}, \cite{Lewicki2000},
\cite{Tropp2004}, \cite{OlshausenCW2009}. The basic assumption underlying
this technique is that a natural signal can be approximated by the
combination of only a small number of elementary components, called
\textit{codewords} or \textit{atoms}, that are chosen from a dictionary
(i.e., the whole collection of all the codewords). Sparse representations
have found successful applications in data interpretation \cite{OlshausenF1996},
\cite{TosicF2011}, source separation \cite{Zibulevsky2001}, \cite{Gribonval2002},
\cite{XuW2011}, signal denoising \cite{AharonEB2006}, \cite{Jafari2011},
coding \cite{SchmidSaugeon2004}, \cite{Kokiopoulou2008}, \cite{Plumbley2009},
classification \cite{Huang2007}, \cite{Mairal2008}, \cite{Schnass2010},
recognition \cite{Wright2009}, impainting \cite{Cevher2009}, \cite{Adler2011}
and many more (see e.g. \cite{Baraniuk2010}).

Two related problems have been studied either separately or jointly
in sparse representations. The first one is sparse coding, that is,
to find the sparse linear decompositions of a signal for a given dictionary.
Efforts dedicated to this problem have resulted in the creation of
a number of algorithms including basis pursuit (BP) \cite{ChenDS1999},
matching pursuit (MP) \cite{MallatZ1993}, orthogonal matching pursuit
(OMP) \cite{Pati1993,Tropp2007}, subspace pursuit (SP) \cite{Dai2008:IT:SubspacePursuit,Tropp2008_CoSamp},
regression shrinkage and selection (LASSO) \cite{Tibshirani1996},
focal under-determined system solver (FOCUSS) \cite{GorodnitskyR1997},
and gradient pursuit (GP) \cite{Blumensath2008}. Sparse decompositions
of a signal, however, rely highly on the degree of fitting between
the data and the dictionary, which leads to the second problem, i.e.
the issue of dictionary design.

An over-complete dictionary, one in which the number of codewords
is greater than the dimension of the signal, can be obtained by either
an analytical or a learning-based approach. The analytical approach
generates the dictionary based on a predefined mathematical transform,
such as discrete Fourier transform (DFT), discrete cosine transform
(DCT), wavelets \cite{Kingsbury2005}, curvelets \cite{Candes1999},
contourlets \cite{DoVetterli2005}, and bandelets \cite{LePennec2005}.
Such dictionaries are relatively easier to obtain and more suitable
for generic signals. In learning-based approaches, however, the dictionaries
are adapted from a set of training data \cite{OlshausenF1996}, \cite{OlshausenF1997},
\cite{EnganAH1999}, \cite{KreutzDelgado2003}, \cite{WipfR2004},
\cite{AharonEB2006}, \cite{Plumbley2007}, \cite{Jarafi2009}, \cite{Gribonval2010},
\cite{Geng2011}. Although this may involve higher computational complexity,
learned dictionaries have the potential to offer improved performance
as compared with predefined dictionaries, since the atoms are derived
to capture the salient information directly from the signals.

Dictionary learning algorithms are often established on an optimization
process involving the iteration between two stages: sparse approximation
and dictionary update. First an initial dictionary is given and a
signal is decomposed as a linear combination of only a few atoms from
the initial dictionary. Then the atoms of the dictionary are trained
with fixed or sometimes unfixed weighting coefficients. After that,
the trained dictionary is used to compute the new weighting coefficients.
The process is iterated until the most suitable dictionary is eventually
obtained.

One of the early algorithms that adopted such a two-step structure
was proposed by Olshausen and Field \cite{OlshausenF1996}, \cite{OlshausenF1997},
where a maximum likelihood (ML) learning method was used to sparsely
code the natural images upon a redundant dictionary. The sparse approximation
step in the ML algorithm \cite{OlshausenF1996} which involves probabilistic
inference is computationally expensive. In a similar probabilistic
framework, Kreutz-Delgado \emph{et al.} \cite{KreutzDelgado2003}
proposed a maximum a posteriori (MAP) dictionary learning algorithm,
where the maximization of the likelihood function as used in \cite{OlshausenF1996}
is replaced by the maximization of posterior probability that a given
signal can be synthesized by a dictionary and the sparse coefficients.
Based on the same ML objective function as in \cite{OlshausenF1996},
Engan \emph{et al.} \cite{EnganAH1999} developed a more efficient
algorithm, called the method of optimal directions (MOD), in which
a closed-form solution for the dictionary update has been proposed.
This method is one of the earliest methods that implements the concept
of sparification process \cite{Rubinstein2010}. Several variants
of this algorithm, such as the iterative least squares (ILS) method,
have also been developed which were summarized in \cite{Engan2007}.
A recursive least squares (RLS) dictionary learning algorithm was
recently presented in \cite{Skretting2010} where the dictionary is
continuously updated as each training vector is being processed, which
is different from the ILS dictionary learning method. Aharon, Elad
and Bruckstein developed the K-SVD algorithm in \cite{AharonEB2006}
by generalizing the K-means algorithm for dictionary learning. This
algorithm uses a similar block-relaxation approach to MOD, but updates
the dictionary on an atom-by-atom basis, without having to compute
matrix inversion as required in the original MOD algorithm. The majorization
method was proposed by \cite{YaghoobiBD2009} in which the original
objective function is substituted by a surrogate function in each
step of the optimization process.

In contrast to the generic dictionaries described above, learning
structure-oriented parametric dictionaries has also attracted attention.
For example, a Gammatone generating function has been used by Yaghoobi
\emph{et al.} \cite{YaghoobiDD2009} to learn dictionaries from audio
data. In \cite{Sallee2003}, a pyramidal wavelet-like transform was
proposed to learn a multiscale structure in the dictionary. Other
constraints have also been considered in the learning process to favor
the desired structures of the dictionaries, such as the translation-invariant
or shift-invariant characteristics of the atoms imposed in \cite{Cadieu2008},
\cite{Blumensath2006}, \cite{Mailhe2008}, \cite{Aharon2008}, \cite{Smaragdis2008}
and the orthogonality between subspaces enforced in \cite{Gribonval2003},
and the de-correlation between the atoms promoted in \cite{Jost2006}.
An advantage of a parametric dictionary lies in its potential for
reducing the number of free parameters and thereby leading to a more
efficient implementation and better convergence of dictionary learning
algorithms \cite{Rubinstein2010}. Other recent efforts in dictionary
learning include the search for robust and computationally efficient
algorithms, such as \cite{Labusch2011}, \cite{Mairal2010}, and \cite{Jafari2011},
and learning dictionaries from multimodal data \cite{Monaci2009},
\cite{Llagostera2010}. Comprehensive reviews of dictionary learning
algorithms can be found in recent survey papers e.g. \cite{Rubinstein2010}
and \cite{Tosic2011}.

In this paper, similar to MOD and K-SVD methods, we focus on the dictionary
update step for generic dictionary learning. We propose a novel optimization
framework where the dictionary update problem is formulated as an
optimization problem on manifolds. The proposed optimization framework
has the following advantages. 
\begin{itemize}
\item In our framework, an \emph{arbitrary} subset of the codewords are
allowed to be updated simultaneously, hence the term \emph{simultaneous
codeword optimization (SimCO)}. This framework can be viewed as a
generalization of the MOD and K-SVD methods: when sparse coefficients
are fixed, the underlying optimization problem is similar to that
in the MOD algorithm; when only one codeword is selected for update,
the optimization problems that arise in both SimCO and K-SVD are identical. 
\item Our framework naturally accommodates a regularization term, motivated
by the ill-condition problem that arises in MOD, K-SVD and primitive
SimCO (detailed in Section \ref{sec:Regularized-SimCO}). We refer
to SimCO with the regularization term as regularized SimCO, which
mitigates the ill-condition problem and hence achieves much better
performance according to our numerical simulations. Note however that
it is not straightforward to extend MOD or K-SVD to the regularized
case. 
\item Though our implementation is based on a simple gradient descent mechanism,
our empirical tests show that the regularized SimCO that updates all
codewords simultaneously enjoys good learning performance and fast
running speed. 
\end{itemize}
Furthermore, we rigorously show that when only one codeword is updated
in each step, the primitive SimCO and K-SVD share the same learning
performance with probability one. As a byproduct, for the first time,
we prove that a gradient search on the Grassmann manifold solves the
rank-one matrix approximation problem with probability one.

\vspace{0cm}

The remainder of the paper is organized as follows. Section \ref{sec:Dictionary-Update-Formulation}
introduces the proposed optimization formulation for dictionary update.
Section \ref{sec:Preliminaries-on-Manifolds} provides necessary preliminaries
on manifolds and shows that dictionary update can be cast as an optimization
problem on manifolds. The implementation details for primitive and
regularized SimCOs are presented in Sections \ref{sec:Dictionary-Update}
and \ref{sec:Regularized-SimCO}, respectively. In Section \ref{sec:Analysis-SimCO},
we rigorously prove the close connection between SimCO and K-SVD.
Numerical results of SimCO algorithms are presented in Section \ref{sec:Simulations}.
Finally, the paper is concluded in Section \ref{sec:Conclusions}.

\section{\label{sec:Dictionary-Update-Formulation}The Optimization Framework
of SimCO}

Dictionary learning is a process of which the purpose is to find an
over-complete dictionary that best represents the training signals.
More precisely, let $\bm{Y}\in\mathbb{R}^{m\times n}$ be the training
data, where each column of $\bm{Y}$ corresponds to one training sample.
For a given dictionary size $d\in\mathbb{Z}^{+}$, the optimal dictionary
$\bm{D}^{*}\in\mathbb{R}^{m\times d}$ is the one that corresponds
to $\inf_{\bm{D}\in\mathbb{R}^{m\times d},\;\bm{X}\in\mathbb{R}^{d\times n}}\;\left\Vert \bm{Y}-\bm{D}\bm{X}\right\Vert _{F}^{2}$,
where $\left\Vert \cdot\right\Vert _{F}$ is the Frobenius norm. Here,
the $i^{th}$ column of $\bm{D}$ is often referred to as the $i^{th}$
\emph{codeword} in the dictionary. In practice, it is typical that
$m<d<n$, i.e., an over-complete dictionary is considered and the
number of training samples is larger than the number of codewords.
Generally speaking, the optimization problem is ill-posed unless extra
constraints are imposed on the dictionary $\bm{D}$ and the coefficient
matrix $\bm{X}$. The most common constraint on $\bm{X}$ is that
$\bm{X}$ is sparse, i.e., the number of nonzero entries in $\bm{X}$,
compared with the total number of entries, is small. 

Most dictionary learning algorithms consist of two stages: sparse
coding and dictionary update. See Algorithm \ref{alg:dictionary-learning}
for the diagram of a typical dictionary learning procedure. In the
sparse coding stage, the goal is to find a sparse $\bm{X}$ to minimize
$\left\Vert \bm{Y}-\bm{D}\bm{X}\right\Vert _{F}^{2}$ for a given
dictionary $\bm{D}$. In practice, the sparse coding problem is often
approximately solved by using either $\ell_{1}$-minimization \cite{Candes2005:DecodingLinearProgramming}
or greedy algorithms, for example, OMP \cite{Tropp2007} and SP \cite{Dai2008:IT:SubspacePursuit}
algorithms.

\begin{algorithm}
\textbf{Task}: find the best dictionary to represent the data sample
matrix $\bm{Y}$.

\textbf{Initialization}: Set the initial dictionary $\bm{D}^{\left(1\right)}$.
Set $J=1$.

\textbf{Repeat} until convergence (use stop rule): 
\begin{itemize}
\item Sparse coding stage: Fix the dictionary $\bm{D}^{\left(J\right)}$
and update $\bm{X}^{\left(J\right)}$ using some sparse coding technique. 
\item Dictionary update stage: Update $\bm{D}^{\left(J\right)}$, and $\bm{X}^{\left(J\right)}$
as appropriate. 
\item $J=J+1$. 
\end{itemize}
\caption{\label{alg:dictionary-learning}A typical dictionary learning algorithm}
\end{algorithm}

The focus of this paper is on the dictionary update stage. There are
different formulations for this stage, leading to substantially different
algorithms. In the MOD \cite{EnganAH1999} method, one fixes the sparse
coding matrix $\bm{X}$ and searches for the optimal dictionary $\bm{D}$,
and hence essentially solves a least squares problem.%
\footnote{When there are no constraints on the norm of the columns of $\bm{D}$,
minimizing $\left\Vert \bm{Y}-\bm{D}\bm{X}\right\Vert _{F}^{2}$ for
given $\bm{Y}$ and $\bm{X}$ is a standard least squares problem
and admits a closed-form solution. When extra constraints on the column
norm are imposed, as we shall show shortly, the optimization problem
is a least squares problem on a product of manifolds. No closed-form
solution has been found. %
} By contrast, in the approach represented by the K-SVD method, one
updates both the dictionary $\bm{D}$ and the nonzero coefficients
in $\bm{X}$. In particular, in each step of the dictionary update
stage of the K-SVD algorithm, one updates \emph{one codeword} of the
dictionary $\bm{D}$ and the nonzero coefficients in the corresponding
row of the matrix $\bm{X}$. After sequentially updating all the codewords
and their corresponding coefficients, the only element fixed is the
sparsity pattern, that is, the locations of the non-zeros in $\bm{X}$.
As has been demonstrated empirically in \cite{AharonEB2006}, the
K-SVD algorithm often enjoys faster convergence and produces a more
accurate dictionary when compared with the MOD method.

The key characteristic of our approach is to update all codewords
and the corresponding non-zero coefficients \emph{simultaneously}.
In our formulation, we assume that the dictionary matrix $\bm{D}$
contains unit $\ell_{2}$-norm columns and the sparsity pattern of
$\bm{X}$ remains unchanged. More specifically, define 
\begin{equation}
\mathcal{D}=\left\{ \bm{D}\in\mathbb{R}^{m\times d}:\;\left\Vert \bm{D}_{:,i}\right\Vert _{2}=1,\;\forall i\in\left[d\right]\right\} ,\label{eq:Feasible-D}
\end{equation}
where $\left\Vert \cdot\right\Vert _{2}$ is the $\ell_{2}$-norm
and the set $\left[d\right]=\left\{ 1,2,\cdots,d\right\} $. The \emph{sparsity
pattern} of $\bm{X}$ is represented by the set $\Omega\subset\left[d\right]\times\left[n\right]$
which contains the indices of all the non-zero entries in $\bm{X}$:
that is, $X_{i,j}\ne0$ for all $\left(i,j\right)\in\Omega$ and $X_{i,j}=0$
for all $\left(i,j\right)\notin\Omega$. Define 
\begin{equation}
\mathcal{X}\left(\Omega\right)=\left\{ \bm{X}\in\mathbb{R}^{d\times n}:\; X_{i,j}=0\;\forall\left(i,j\right)\notin\Omega\right\} .\label{eq:Feasible-X}
\end{equation}
The dictionary update problem under consideration is given by 
\begin{equation}
\underset{\bm{D}\in\mathcal{D}}{\inf}\;\underset{\bm{X}\in\mathcal{X}\left(\Omega\right)}{\inf}\;\left\Vert \bm{Y}-\bm{D}\bm{X}\right\Vert _{F}^{2}.\label{eq:Primitive-SimCO}
\end{equation}
Note that the optimal $\bm{X}$ that minimizes $\left\Vert \bm{Y}-\bm{D}\bm{X}\right\Vert _{F}^{2}$
varies as $\bm{D}$ changes. An update in $\bm{D}$ implies an update
of the corresponding optimal $\bm{X}$. Hence, both $\bm{D}$ and
$\bm{X}$ are simultaneously updated. We refer to this optimization
framework as \emph{primitive SimCO}. 

Another optimization framework proposed in this paper is the so called
\emph{regularized SimCO}. The related optimization problem is given
by 
\begin{equation}
\underset{\bm{D}\in\mathcal{D}}{\inf}\;\underset{\bm{X}\in\mathcal{X}\left(\Omega\right)}{\inf}\;\left\Vert \bm{Y}-\bm{D}\bm{X}\right\Vert _{F}^{2}+\mu\left\Vert \bm{X}\right\Vert _{F}^{2},\label{eq:Regularized-SimCO}
\end{equation}
where $\mu>0$ is a properly chosen constant. The motivation of introducing
the regularization term $\mu\left\Vert \bm{X}\right\Vert _{F}^{2}$
is presented in Section \ref{sec:Regularized-SimCO}. 

The ideas of SimCO can be generalized: instead of updating all codewords
simultaneously, one can update an arbitrary subset of codewords and
the corresponding coefficients. More precisely, let $\mathcal{I}\subseteq\left[d\right]$
be the index set of the codewords to be updated. That is, only codewords
$\bm{D}_{:,i}$'s, $i\in\mathcal{I}$, are to be updated while all
other codewords $\bm{D}_{:,j}$'s, $j\notin\mathcal{I}$, remain constant.
Let $\bm{D}_{:,\mathcal{I}}$ denote the sub-matrix of $\bm{D}$ formed
by the columns of $\bm{D}$ indexed by $\mathcal{I}$. Let $\bm{X}_{\mathcal{I},:}$
denote the sub-matrix of $\bm{X}$ consisting of the rows of $\bm{X}$
indexed by $\mathcal{I}$. Define 
\[
\bm{Y}_{r}=\bm{Y}-\bm{D}_{:,\mathcal{I}^{c}}\bm{X}_{\mathcal{I}^{c},:},
\]
where $\mathcal{I}^{c}$ is a set complementary to $\mathcal{I}$.
Then $\bm{Y}-\bm{D}\bm{X}=\bm{Y}_{r}-\bm{D}_{:,\mathcal{I}}\bm{X}_{\mathcal{I},:}$.
Then the optimization problems in SimCO can be written as 
\[
\underset{\bm{D}_{:,\mathcal{I}}:\;\bm{D}\in\mathcal{D}}{\inf}\; f_{\mathcal{I}}\left(\bm{D}\right),
\]
where the objective function $f_{\mathcal{I}}\left(\bm{D}\right)$
is given by
\begin{equation}
f_{\mathcal{I}}\left(\bm{D}\right)=\underset{\bm{X}_{\mathcal{I},:}:\;\bm{X}\in\mathcal{X}\left(\Omega\right)}{\inf}\;\left\Vert \bm{Y}_{r}-\bm{D}_{:,\mathcal{I}}\bm{X}_{\mathcal{I},:}\right\Vert _{F}^{2}\label{eq:primitive-SimCO-f}
\end{equation}
 for primitive SimCO and 
\begin{equation}
f_{\mathcal{I}}\left(\bm{D}\right)=\underset{\bm{X}_{\mathcal{I},:}:\;\bm{X}\in\mathcal{X}\left(\Omega\right)}{\inf}\;\left\Vert \bm{Y}_{r}-\bm{D}_{:,\mathcal{I}}\bm{X}_{\mathcal{I},:}\right\Vert _{F}^{2}+\mu\left\Vert \bm{X}_{\mathcal{I},:}\right\Vert _{F}^{2}\label{eq:regularized-SimCO-f}
\end{equation}
for regularized SimCO, respectively. The algorithmic details for solving
primitive and regularized SimCO are presented in Sections \ref{sec:Dictionary-Update}
and \ref{sec:Regularized-SimCO} respectively. 

The connection between our formulation and those in MOD and K-SVD
is clear. When sparse coefficients are fixed, the underlying optimization
problem is similar to that in MOD. When only one codeword is selected
for update, the formulation in (\ref{eq:primitive-SimCO-f}) is identical
to the optimization formulation treated in K-SVD. 

There are also fundamental differences between our framework and those
in MOD and K-SVD. Compared with MOD, our formulation puts a constraint
(\ref{eq:Feasible-D}) on the $\ell_{2}$-norm of the columns of the
dictionary matrix. This constraint is motivated by the following reasons.
\begin{enumerate}
\item The performance of a given dictionary is invariant to the column norms.
The performance of a given dictionary $\bm{D}$ is described by how
the product $\bm{D}\bm{X}$ approximates the training samples $\bm{Y}$.
By scaling the corresponding rows in $\bm{X}$, one can keep the product
$\bm{D}\bm{X}$ invariant to any nonzero scaling of the columns in
$\bm{D}$. 
\item A normalized dictionary $\bm{D}\in\mathcal{D}$ is preferred in the
sparse coding stage. Sparse coding algorithms rely heavily on the
magnitudes of the coefficients $X_{i,j}$'s, $\left(i,j\right)\in\left[d\right]\times\left[n\right]$,
which are affected by the column norms of $\bm{D}$. It is a standard
practice to normalize the columns of $\bm{D}$ before applying sparse
coding algorithms. 
\item A normalized dictionary $\bm{D}\in\mathcal{D}$ is required in regularized
SimCO. The regularization term $\mu\left\Vert \bm{X}\right\Vert _{F}^{2}$
is useful only when the column norms of $\bm{D}$ are fixed. To see
this, let $\bm{D}_{1},\bm{D}_{2}\in\mathbb{R}^{m\times d}$ be two
dictionaries whose columns are only different in scaling; it can be
shown that in this case the optimal $\bm{X}$ for the minimization
of $\left\Vert \bm{Y}-\bm{D}\bm{X}\right\Vert _{F}^{2}+\mu\left\Vert \bm{X}\right\Vert _{F}^{2}$
can be very different  and so is the regularization term.
\end{enumerate}
More subtly, the singularity phenomenon that motivates regularized
SimCO depends upon the normalized columns. This point will be detailed
in Section \ref{sec:Regularized-SimCO}.

Our formulation naturally accommodates an inclusion of the regularization
term in (\ref{eq:Regularized-SimCO}). As will be shown in Sections
\ref{sec:Regularized-SimCO} and \ref{sec:Simulations}, the regularization
term improves the learning performance significantly. Note that it
is not clear how to extend MOD or K-SVD for the regularized case.
In the dictionary update step of MOD, the coefficient matrix $\bm{X}$
is fixed. The regularization term becomes a constant and does not
appear in the optimization problem. The main idea of K-SVD is to use
SVD to solve the corresponding optimization problem. However, it is
not clear how to employ SVD to solve the regularized optimization
problem in (\ref{eq:regularized-SimCO-f}) when $\left|\mathcal{I}\right|=1$.

\section{\label{sec:Preliminaries-on-Manifolds}Preliminaries on Manifolds}

Our approach for solving the optimization problem (\ref{eq:Primitive-SimCO})
relies on the notion of Stiefel and Grassmann manifolds. In particular,
the Stiefel manifold $\mathcal{U}_{m,1}$ is defined as $\mathcal{U}_{m,1}=\left\{ \bm{u}\in\mathbb{R}^{m}:\;\bm{u}^{T}\bm{u}=1\right\} .$
The Grassmann manifold $\mathcal{G}_{m,1}$ is defined as $\mathcal{G}_{m,1}=\left\{ \mbox{span}\left(\bm{u}\right):\;\bm{u}\in\mathcal{U}_{m,1}\right\} .$
Here, the notations $\mathcal{U}_{m,1}$ and $\mathcal{G}_{m,1}$
follow from the convention in \cite{Edelman:1999:Algorithms-Orthogonality-Constraints,Dai2011:LRMCGeometric}.
Note that each element in $\mathcal{U}_{m,1}$ is a unit-norm vector
while each element in $\mathcal{G}_{m,1}$ is a one-dimensional subspace
in $\mathbb{R}^{m}$. For any given $\bm{u}\in\mathcal{U}_{m,1}$,
it can generate a one-dimensional subspace $\mathscr{U}\in\mathcal{G}_{m,1}$.
Meanwhile, any given $\mathscr{U}\in\mathcal{G}_{m,1}$ can be generated
from different $\bm{u}\in\mathcal{U}_{m,1}$: if $\mathscr{U}=\mbox{span}\left(\bm{u}\right)$,
then $\mathscr{U}=\mbox{span}\left(-\bm{u}\right)$ as well.

With these definitions, the dictionary $\bm{D}$ can be interpreted
as the Cartesian product of $d$ many Stiefel manifolds $\mathcal{U}_{m,1}$.
Each codeword (column) in $\bm{D}$ is one element in $\mathcal{U}_{m,1}$.
It looks straightforward that optimization over $\bm{D}$ is an optimization
over the product of Stiefel manifolds.

What is not so obvious is that the optimization is actually over the
product of Grassmann manifolds. For any given pair $\left(\bm{D},\bm{X}\right)$,
if the signs of $\bm{D}_{:,i}$ and $\bm{X}_{i,:}$ change simultaneously,
the value of the objective function $\left\Vert \bm{Y}-\bm{D}\bm{X}\right\Vert _{F}^{2}$
stays the same. Let $\bm{D}=\left[\bm{D}_{:,1},\cdots,\bm{D}_{:,i-1},\bm{D}_{:,i},\bm{D}_{:,i+1},\cdots,\bm{D}_{:,d}\right]$
and $\bm{D}^{\prime}=\left[\bm{D}_{:,1},\cdots,\bm{D}_{:,i-1},-\bm{D}_{:,i},\bm{D}_{:,i+1},\cdots,\bm{D}_{:,d}\right]$.
Then it is straightforward to verify that $f_{\left[d\right]}\left(\bm{D}\right)=f_{\left[d\right]}\left(\bm{D}^{\prime}\right)$.
In other words, it does not matter what $\bm{D}_{:,i}$ is; what matters
is the generated subspace $\mbox{span}\left(\bm{D}_{:,i}\right)$.
As shall become explicit later, this phenomenon has significant impacts
on algorithm design and analysis. 

It is worth noting that the performance of a given dictionary is invariant
to the permutations of the codewords. However, how to effectively
address this permutation invariance analytically and algorithmically
remains an open problem.

\section{\label{sec:Dictionary-Update}Implementation Details for Primitive
SimCO}

This section presents the algorithmic details of primitive SimCO.
For proof-of-concept, we use a simple gradient descent method. The
gradient computation is detailed in Subsection \ref{sub:Gradient-computation}.
How to search on the manifold product space is specified in Subsection
\ref{sub:Line-search}. The overall procedure for dictionary update
is described in Algorithm \ref{alg:update-dictionary}. Note that
one may apply second-order optimization methods, for example, the
trust region method \cite{Absil2008:Book:OptimizatoinManifolds},
for SimCO. The convergence rate is expected to be much faster than
that of gradient descent methods. However, this is beyond the scope
of this paper.

\subsection{\label{sub:Gradient-computation}Gradient computation }

In this subsection, we compute the $f_{\mathcal{I}}\left(\bm{D}\right)$
in (\ref{eq:primitive-SimCO-f}) and the corresponding gradient $\nabla f_{\mathcal{I}}\left(\bm{D}\right)$. 

The computation of $f_{\mathcal{I}}\left(\bm{D}\right)$ involves
solving the corresponding least squares problem. For a given $j\in\left[n\right]$,
let $\Omega\left(:,j\right)=\left\{ i:\;\left(i,j\right)\in\Omega\right\} $.
Similarly, we define $\Omega\left(i,:\right)=\left\{ j:\;\left(i,j\right)\in\Omega\right\} $.
Let $\bm{X}_{\mathcal{I}\cap\Omega\left(:,j\right),j}$ be the sub-vector
of $\bm{X}_{:,j}$ indexed by $\mathcal{I}\cap\Omega\left(:,j\right)$,
and $\bm{D}_{:,\mathcal{I}\cap\Omega\left(:,j\right)}$ be the sub-matrix
of $\bm{D}$ composed on the columns indexed by $\mathcal{I}\cap\Omega\left(:,j\right)$.
It is straightforward to verify that 
\[
\left\Vert \bm{Y}_{r}-\bm{D}_{:,\mathcal{I}}\bm{X}_{\mathcal{I},:}\right\Vert _{F}^{2}=\sum_{j=1}^{n}\left\Vert \left(\bm{Y}_{r}\right)_{:,j}-\bm{D}_{:,\mathcal{I}\cap\Omega\left(:,j\right)}\bm{X}_{\mathcal{I}\cap\Omega\left(:,j\right),j}\right\Vert _{2}^{2},
\]
and
\begin{equation}
f_{\mathcal{I}}\left(\bm{D}\right)=\sum_{j=1}^{n}\;\underbrace{\underset{\bm{X}_{\mathcal{I}\cap\Omega\left(:,j\right),j}}{\inf}\;\left\Vert \left(\bm{Y}_{r}\right)_{:,j}-\bm{D}_{:,\mathcal{I}\cap\Omega\left(:,j\right)}\bm{X}_{\mathcal{I}\cap\Omega\left(:,j\right),j}\right\Vert _{2}^{2}}_{f_{\mathcal{I},j}\left(\bm{D}\right)}.\label{eq:f-I-j-D}
\end{equation}
 Note that every atomic function $f_{\mathcal{I},j}\left(\bm{D}\right)$
corresponds to a least squares problem of the form $\inf_{\bm{x}}\;\left\Vert \bm{y}-\bm{A}\bm{x}\right\Vert _{F}^{2}$.
The optimal $\bm{X}^{*}$ admits the following closed-from 
\begin{align}
 & \bm{X}_{i,j}^{*}=0,\;\forall\left(i,j\right)\notin\Omega,\;\bm{X}_{\mathcal{I}^{c},:}^{*}=\bm{X}_{\mathcal{I}^{c},:}\nonumber \\
 & \bm{X}_{\mathcal{I}\cap\Omega\left(:,j\right),j}^{*}=\bm{D}_{:,\mathcal{I}\cap\Omega\left(:,j\right)}^{\dagger}\left(\bm{Y}_{r}\right)_{:,j},\;\forall j\in\left[n\right],\label{eq:optimal-X}
\end{align}
 where the superscript $\dagger$ denotes the pseudo-inverse of a
matrix. In practice, $\bm{X}_{\mathcal{I}\cap\Omega\left(:,j\right),j}^{*}$
can be computed via low complexity methods, for example, the conjugate
gradient method \cite{Jorge2006:Numerical-Optimization}, to avoid
the more computationally expensive pseudo-inverse. 

The gradient of $f_{\mathcal{I}}\left(\bm{D}\right)$ is computed
as follows. Let us consider a general least squares problem $f_{LS}\left(\bm{A}\right)=\inf_{\bm{x}}\;\left\Vert \bm{y}-\bm{A}\bm{x}\right\Vert _{2}^{2}$.
Clearly the optimal $\bm{x}^{*}=\bm{A}^{\dagger}\bm{y}$ is a function
of $\bm{A}$. With slight abuse in notations, write $f_{LS}\left(\bm{A}\right)$
as $f_{LS}\left(\bm{A},\bm{x}^{*}\right)$. Then 
\begin{align}
\nabla_{\bm{A}}f_{LS} & =\frac{\partial f_{LS}\left(\bm{A},\bm{x}^{*}\right)}{\partial\bm{A}}+\frac{\partial f_{LS}\left(\bm{A},\bm{x}^{*}\right)}{\partial\bm{x}^{*}}\cdot\frac{d\bm{x}^{*}}{d\bm{A}}\nonumber \\
 & =-2\left(\bm{y}-\bm{A}\bm{x}^{*}\right)\bm{x}^{*T}+\bm{0}\frac{d\bm{x}^{*}}{d\bm{A}}\nonumber \\
 & =-2\left(\bm{y}-\bm{A}\bm{x}^{*}\right)\bm{x}^{*T},\label{eq:gradient-quadratic}
\end{align}
where the second equality holds because $\bm{x}^{*}$ minimizes $\left\Vert \bm{y}-\bm{A}\bm{x}^{*}\right\Vert _{2}^{2}$
and hence $\frac{\partial f}{\partial\bm{x}^{*}}=\bm{0}$. Based on
(\ref{eq:gradient-quadratic}), the gradient of $f_{\mathcal{I}}\left(\bm{D}\right)$,
with respect to $\bm{D}_{:,i}$, $i\in\mathcal{I}$, can be computed
via 
\begin{align}
\nabla_{\bm{D}_{:,i}}f_{\mathcal{I}}\left(\bm{D}\right) & =-2\left(\bm{Y}-\bm{D}\bm{X}^{*}\right)_{:,\Omega\left(i,:\right)}\bm{X}_{i,\Omega\left(i,:\right)}^{*T}\nonumber \\
 & =-2\left(\bm{Y}-\bm{D}\bm{X}^{*}\right)\bm{X}_{i,:}^{*T}.\label{eq:gradient}
\end{align}
 Here, $\Omega\left(i,:\right)$ gives the columns of $\bm{Y}$ whose
sparse representation involves the codeword $\bm{D}_{:,i}$.

When $\mathcal{I}=\left[d\right]$, the formulas for $\bm{X}^{*}$
and $\nabla_{\bm{D}}f$ can be simplified to 
\begin{align*}
 & \bm{X}_{i,j}^{*}=0,\;\forall\left(i,j\right)\notin\Omega,\\
 & \bm{X}_{\Omega\left(:,j\right),j}^{*}=\bm{D}_{:,\Omega\left(:,j\right)}^{\dagger}\bm{Y}_{:,j},\;\forall j\in\left[n\right],\;\mbox{and}\;\\
 & \nabla_{\bm{D}}f_{\left[d\right]}\left(\bm{D}\right)=-2\left(\bm{Y}-\bm{D}\bm{X}^{*}\right)\bm{X}^{*T}.
\end{align*}

\subsection{\label{sub:Line-search}Line search along the gradient descent direction}

The line search mechanism used in this paper is significantly different
from the standard one for the Euclidean space. In a standard line
search algorithm, the $k^{th}$ iteration outputs an updated variable
$\bm{x}^{\left(k\right)}$ via 
\begin{equation}
\bm{x}^{\left(k\right)}=\bm{x}^{\left(k-1\right)}-t\cdot\nabla_{\bm{x}}f\left(\bm{x}^{\left(k-1\right)}\right),\label{eq:line-search-normal}
\end{equation}
 where $f\left(\bm{x}\right)$ is the objective function to be minimized,
and $t\in\mathbb{R}^{+}$ is a properly chosen step size. However,
a direct application of (\ref{eq:line-search-normal}) may result
in a dictionary $\bm{D}\notin\mathcal{D}$.

The line search path in this paper is restricted to the product of
Grassmann manifolds. This is because, as has been discussed in Section
\ref{sec:Preliminaries-on-Manifolds}, the objective function $f_{\mathcal{I}}$
is indeed a function on the product of Grassmann manifolds. On the
Grassmann manifold $\mathcal{G}_{m,1}$, the geodesic path plays the
same role as the straight line in the Euclidean space: given any two
distinct points on $\mathcal{G}_{m,1}$, the shortest path that connects
these two points is geodesic \cite{Edelman:1999:Algorithms-Orthogonality-Constraints}.
In particular, let $\mathscr{U}\in\mathcal{G}_{m,1}$ be a one-dimensional
subspace and $\bm{u}\in\mathcal{U}_{m,1}$ be the corresponding generator
matrix (not unique).%
\footnote{The generator matrix $\bm{u}$ is a vector in this case.%
} Consider a search direction $\bm{h}\in\mathbb{R}^{m}$ with $\left\Vert \bm{h}\right\Vert _{2}=1$
and $\bm{h}^{T}\bm{u}=0$. Then the geodesic path starting from $\bm{u}$
along the direction $\bm{h}$ is given by \cite{Edelman:1999:Algorithms-Orthogonality-Constraints}
\[
\bm{u}\left(t\right)=\bm{u}\cdot\cos t+\bm{h}\cdot\sin t,\; t\in\mathbb{R}.
\]
Note that $\bm{u}\left(t\right)=-\bm{u}\left(t+\pi\right)$ and hence
$\mbox{span}\left(\bm{u}\left(t\right)\right)=\mbox{span}\left(\bm{u}\left(t+\pi\right)\right)$.
In practice, one can restrict the search path within the interval
$t\in\left[0,\pi\right)$. 

For the dictionary update problem at hand, the line search path is
defined as follows. Let $\bm{g}_{i}=\nabla_{\bm{D}_{:,i}}f_{\mathcal{I}}\left(\bm{D}\right)$
be the gradient vector defined in (\ref{eq:gradient}). We define
\begin{equation}
\bar{\bm{g}}_{i}=\bm{g}_{i}-\bm{D}_{:,i}\bm{D}_{:,i}^{T}\bm{g}_{i},\;\forall i\in\mathcal{I},\label{eq:gradient-Grassmann-manifold}
\end{equation}
so that $\bar{\bm{g}}_{i}$ and $\bm{D}_{:,i}$ are orthogonal. The
line search path for dictionary update, say $\bm{D}\left(t\right)$,
$t\ge0$, is given by \cite{Edelman:1999:Algorithms-Orthogonality-Constraints}
\begin{equation}
\left\{ \begin{array}{l}
\bm{D}_{:,i}\left(t\right)=\bm{D}_{:,i}\quad{\rm if}\; i\notin\mathcal{I}\;{\rm or}\;\left\Vert \bar{\bm{g}}_{i}\right\Vert _{2}=0,\\
\bm{D}_{:,i}\left(t\right)=\bm{D}_{:,i}\cos\left(\left\Vert \bar{\bm{g}}_{i}\right\Vert _{2}t\right)-\left(\bar{\bm{g}}_{i}/\left\Vert \bar{\bm{g}}_{i}\right\Vert _{2}\right)\sin\left(\left\Vert \bar{\bm{g}}_{i}\right\Vert _{2}t\right)\\
\qquad\qquad\qquad\quad{\rm if}\; i\in\mathcal{I}\;{\rm and}\;\left\Vert \bar{\bm{g}}_{i}\right\Vert _{2}\ne0.
\end{array}\right.\label{eq:search-path}
\end{equation}

Algorithm \ref{alg:update-dictionary} summarizes one iteration of
the proposed line search algorithm. For proof-of-concept and implementation
convenience, we use the method of golden section search (see \cite{Press2007:NumericalRecipes}
for a detailed description). The idea is to use the golden ratio to
successively narrow the searching range of $t$ inside which a local
minimum exists. To implement this idea, we design a two-step procedure
in Algorithm \ref{alg:update-dictionary}: in the first step (Part
A), we increase/decrease the range of $t$, i.e., $\left(0,t_{4}\right)$,
so that it contains a local minimum and the objective function looks
unimodal in this range; in the second step (Part B), we use the golden
ratio to narrow the range so that we can accurately locate the minimum.
Note that the proposed algorithm is by no means optimized. Other ways
to do a gradient descent efficiently can be found in \cite[Chapter 3]{Jorge2006:Numerical-Optimization}. 

\begin{algorithm}
\caption{\label{alg:update-dictionary}One iteration of the line search algorithm
for dictionary update.}

\textbf{Task}: Use line search mechanism to update the dictionary
$\bm{D}$.

\textbf{Input}: $\bm{Y}$, $\bm{D}$, $\bm{X}$

\textbf{Output}: $\bm{D}^{\prime}$ and $\bm{X}^{\prime}$.

\textbf{Parameters}: $t_{4}>0$: initial step size. $g_{\min}>0$:
the threshold below which a gradient can be viewed as zero.

\textbf{Initialization}: Let $c=\left(\sqrt{5}-1\right)/2$. 
\begin{enumerate}
\item Let $t_{1}=0$. Compute $f\left(\bm{D}\right)$ using (\ref{eq:f-I-j-D})
and the corresponding gradient $\bar{\bm{g}}_{i}$ on the Grassmann
manifold using (\ref{eq:gradient-Grassmann-manifold}) and (\ref{eq:gradient}).
If $\left\Vert \bar{\bm{g}}_{i}\right\Vert _{2}\le g_{\min}\left\Vert \bm{Y}\right\Vert _{F}^{2}$
for all $i\in\mathcal{I}$, then $\bm{D}^{\prime}=\bm{D}$, $\bm{X}^{\prime}=\bm{X}$,
and quit. 
\item Let $t_{3}=ct_{4}$ and $t_{2}=\left(1-c\right)t_{4}$. 
\end{enumerate}
\textbf{Part A}: the goal is to find $t_{4}>0$ s.t. $f\left(\bm{D}\left(t_{1}\right)\right)>f\left(\bm{D}\left(t_{2}\right)\right)>f\left(\bm{D}\left(t_{3}\right)\right)\le f\left(\bm{D}\left(t_{4}\right)\right)$.
Iterate the following steps. 
\begin{enumerate}
\item [3)]\setcounter{enumi}{3}If $f\left(\bm{D}\left(t_{1}\right)\right)\le f\left(\bm{D}\left(t_{2}\right)\right)$,
then $t_{4}=t_{2}$, $t_{3}=ct_{4}$ and $t_{2}=\left(1-c\right)t_{4}$. 
\item Else if $f\left(\bm{D}\left(t_{2}\right)\right)\le f\left(\bm{D}\left(t_{3}\right)\right)$,
then $t_{4}=t_{3}$, $t_{3}=t_{2}$ and $t_{2}=\left(1-c\right)t_{4}$. 
\item Else if $f\left(\bm{D}\left(t_{3}\right)\right)>f\left(\bm{D}\left(t_{4}\right)\right)$,
then $t_{2}=t_{3}$, $t_{3}=t_{4}$ and $t_{4}=t_{3}/c$. 
\item Otherwise, quit the iteration. 
\end{enumerate}
\textbf{Part B}: the goal is to shrink the interval length $t_{4}-t_{1}$
while trying to keep the relation $f\left(\bm{D}\left(t_{1}\right)\right)>f\left(\bm{D}\left(t_{2}\right)\right)>f\left(\bm{D}\left(t_{3}\right)\right)$.
Iterate the following steps until $t_{4}-t_{1}$ is sufficiently small. 
\begin{enumerate}
\item [7)]\setcounter{enumi}{7}If $f\left(\bm{D}\left(t_{1}\right)\right)>f\left(\bm{D}\left(t_{2}\right)\right)>f\left(\bm{D}\left(t_{3}\right)\right)$,
then $t_{1}=t_{2}$, $t_{2}=t_{3}$ and $t_{3}=t_{1}+c\left(t_{4}-t_{1}\right)$. 
\item Else $t_{4}=t_{3}$, $t_{3}=t_{2}$ and $t_{2}=t_{1}+\left(1-c\right)\left(t_{4}-t_{1}\right)$. 
\end{enumerate}
\textbf{Output}: Let $t^{*}=\underset{t\in\left\{ t_{1},t_{2},t_{3},t_{4}\right\} }{\arg\;\min}\; f\left(\bm{D}\left(t\right)\right)$
and $\bm{D}^{\prime}=\bm{D}\left(t^{*}\right)$. Compute $\bm{X}^{\prime}$
according to (\ref{eq:optimal-X}). 
\end{algorithm}

\section{\label{sec:Regularized-SimCO}Implementation Details for Regularized
SimCO}

As will be detailed in Section \ref{sub:ill-condition-case}, MOD,
K-SVD and primitive SimCO may result in ill-conditioned dictionaries.
Regularized SimCO method  (\ref{eq:Regularized-SimCO}) is designed
to mitigate this problem. 

The ill-condition of the dictionary can be described as follows. Fix
the sparsity pattern $\Omega$. The matrix $\bm{D}_{:,\Omega\left(:,j\right)}$
contains the codewords that are involved in representing the training
sample $\bm{Y}_{:,j}$. We say the dictionary $\bm{D}$ is ill-conditioned
with respect to the sparsity pattern $\Omega$ if 
\[
0\approx\lambda_{\min}\left(\bm{D}_{:,\Omega\left(:,j\right)}\right)\ll\lambda_{\max}\left(\bm{D}_{:,\Omega\left(:,j\right)}\right)
\]
 for some $j\in\left[n\right]$. Here, $\lambda_{\min}\left(\cdot\right)$
and $\lambda_{\max}\left(\cdot\right)$ give the smallest and largest
singular values of a matrix, respectively.

The ill-condition of $\bm{D}$ brings two problems: 
\begin{enumerate}
\item Slow convergence in the dictionary update stage. When $\lambda_{\min}\left(\bm{D}_{:,\Omega\left(:,j\right)}\right)$
is close to zero, the curvature (Hessian matrix) of $f_{\mathcal{I}}\left(\bm{D}\right)$
is large. The gradient changes significantly in the neighborhood of
a singular point. Gradient descent algorithms typically suffer from
a very slow convergence rate. 
\item Instability in the subsequent sparse coding stage. When $\lambda_{\min}\left(\bm{D}_{:,\Omega\left(:,j\right)}\right)$
is close to zero, the solution to the least squares problem $\underset{\bm{X}_{\Omega\left(:,j\right),j}}{\inf}\;\left\Vert \bm{Y}_{:,j}-\bm{D}_{:,\Omega\left(:,j\right)}\bm{X}_{\Omega\left(:,j\right),j}\right\Vert _{F}^{2}$
becomes unstable: small changes in $\bm{Y}_{:,j}$ often result in
very different least squares solutions $\bm{X}_{\Omega\left(:,j\right),j}^{*}$.
It is well known that the stability of sparse coding relies on the
so called restricted isometry condition (RIP) \cite{Candes2005:DecodingLinearProgramming},
which requires that the singular values of submatrices of $\bm{D}$
concentrate around $1$. An ill-conditioned $\bm{D}$ violates RIP
and hence results in sparse coefficients that are sensitive to noise. 
\end{enumerate}
\vspace{0cm}

It is worth mentioning that the above discussion on the ill-condition
problem depends upon the unit-norm columns. To see this, consider
a dictionary with orthonormal columns. It is clearly well-conditioned.
However, if one picks a column of the dictionary matrix and scales
it arbitrarily small, the resulted dictionary will then become ill-conditioned.
Hence, a constraint (\ref{eq:Feasible-D}) on column norms is necessary
for the discussion of the condition number of a dictionary. 

It is also worth mentioning the difference between a stationary point
and an ill-conditioned dictionary. In both cases, it is typical that
the objective function stops decreasing as the number of iterations
increases. It is therefore difficult to distinguish these two cases
by looking at the objective function only. However, the difference
becomes apparent by checking the gradient: the gradient is close to
zero in the neighborhood of a stationary point while it becomes large
in the neighborhood of a singular point. This phenomenon is not isolated
as it was also observed in the manifold learning approach for the
low-rank matrix completion problem \cite{Dai2011:LRMCGeometric}.

To mitigate the problem brought by ill-conditioned dictionaries, we
propose regularized SimCO in (\ref{eq:Regularized-SimCO}). Note that
when $\bm{D}$ is ill-conditioned, the optimal $\bm{X}^{*}$ for the
least squares problem in primitive SimCO is typically large. By adding
the regularization term to the objective function, the search path
is {}``pushed'' towards a well-conditioned one. 

Algorithm \ref{alg:update-dictionary} can be directly applied to
regularized SimCO. The only required modifications are the computations
of the new objective function (\ref{eq:regularized-SimCO-f}) and
the corresponding gradient. Similar to primitive SimCO, the objective
function (\ref{eq:regularized-SimCO-f}) in regularized SimCO can
be decomposed into a sum of atomic functions, i.e., 
\begin{align}
f_{\mathcal{I}}\left(\bm{D}\right) & =\sum_{j=1}^{n}\underset{\bm{X}_{\mathcal{I}\cap\Omega\left(:,j\right),j}}{\inf}\left(\left\Vert \left(\bm{Y}_{r}\right)_{:,j}-\bm{D}_{:,\mathcal{I}\cap\Omega\left(:,j\right)}\bm{X}_{\mathcal{I}\cap\Omega\left(:,j\right),j}\right\Vert _{2}^{2}\right.\nonumber \\
 & \qquad\quad\underbrace{\qquad\qquad\qquad\qquad\qquad\left.+\mu\left\Vert \bm{X}_{\mathcal{I}\cap\Omega\left(:,j\right),j}\right\Vert _{2}^{2}\right)}_{f_{\mathcal{I},j}\left(\bm{D}\right)}.\label{eq:f-I-tilde}
\end{align}
One needs to solve the least squares problems in atomic functions
(\ref{eq:f-I-tilde}). Let $m_{j}=\left|\mathcal{I}\cap\Omega\left(:,j\right)\right|$.
It is clear that $\bm{D}_{:,\bm{X}_{\mathcal{I}\cap\Omega\left(:,j\right),j}}\in\mathbb{R}^{m\times m_{j}}$
and $\bm{X}_{\mathcal{I}\cap\Omega\left(:,j\right),j}\in\mathbb{R}^{m_{j}}$.
Define 
\[
\widetilde{\bm{Y}}_{r,j}=\left[\begin{array}{c}
\left(\bm{Y}_{r}\right)_{:,j}\\
\bm{0}_{m_{j}}
\end{array}\right],\;\mbox{and }\widetilde{\bm{D}}_{j}=\left[\begin{array}{c}
\bm{D}_{:,\mathcal{I}\cap\Omega\left(:,j\right)}\\
\sqrt{\mu}\cdot\bm{I}_{m_{j}}
\end{array}\right],
\]
 where $\bm{0}_{m_{j}}$ is the zero vector of length $m_{j}$, and
$\bm{I}_{m_{j}}$ is the $m_{j}\times m_{j}$ identity matrix. The
optimal $\bm{X}_{\mathcal{I}\cap\Omega\left(:,j\right),j}^{*}$ to
solve the least squares problem in (\ref{eq:f-I-tilde}) is given
by 
\begin{equation}
\bm{X}_{\mathcal{I}\cap\Omega\left(:,j\right),j}^{*}=\widetilde{\bm{D}}_{j}^{\dagger}\widetilde{\bm{Y}}_{r,j}.\label{eq:Optimal-X-tilde}
\end{equation}
The corresponding value of the objective function is therefore 
\begin{equation}
f_{\mathcal{I}}\left(\bm{D}\right)=\left\Vert \bm{Y}_{r}-\bm{D}\bm{X}^{*}\right\Vert _{F}^{2}+\mu\cdot\left\Vert \bm{X}_{\mathcal{I},:}^{*}\right\Vert _{F}^{2}.\label{eq:f-I-tilde-computation}
\end{equation}
The gradient computation is similar to that for primitive SimCO. It
can be verified that 
\begin{equation}
\nabla_{\bm{D}_{:,\mathcal{I}}}f_{\mathcal{I}}\left(\bm{D}\right)=-2\left(\bm{Y}-\bm{D}\bm{X}^{*}\right)\bm{X}_{\mathcal{I},:}^{*T}.\label{eq:gradient-tilde}
\end{equation}
Replacing (\ref{eq:f-I-j-D}) and (\ref{eq:gradient}) in Algorithm
\ref{alg:update-dictionary} by (\ref{eq:f-I-tilde-computation})
and (\ref{eq:gradient-tilde}) respectively, we obtain a gradient
descent implementation for regularized SimCO.

In practice, one may consider first using regularized SimCO to obtain
a reasonably good dictionary and then employ primitive SimCO to refine
the dictionary further. This two-step procedure often results in a
well-conditioned dictionary that fits the training data. Please see
the simulation part (Section \ref{sec:Simulations}) for an example.

\section{\label{sec:Analysis-SimCO}Convergence of Primitive SimCO}

The focus of this section is on the convergence performance of  primitive
SimCO when the index set $\mathcal{I}$ contains only one index. The
analysis of this case shows the close connection between primitive
SimCO and K-SVD. More specifically, as we discussed in Section \ref{sec:Dictionary-Update-Formulation},
when $\left|\mathcal{I}\right|=1$, the optimization formulations
of primitive SimCO and K-SVD are exactly the same. To solve this optimization
problem, primitive SimCO uses a gradient descent algorithm while K-SVD
employs singular value decomposition (SVD). In Theorem \ref{thm:Gradient-Works}
of this section, we shall prove that a gradient descent finds a global
optimum with probability one. Hence, when $\left|\mathcal{I}\right|=1$,
the learning performance of primitive SimCO and K-SVD are the same.
Note that, even though the general case when $\left|\mathcal{I}\right|>1$
is more interesting, its convergence is much more difficult to analyze. 

The analysis for the case of $\left|\mathcal{I}\right|=1$ helps in
understanding where the performance gain of SimCO comes from. Theorem
\ref{thm:Gradient-Works} shows the equivalence between K-SVD and
primitive SimCO when $\left|\mathcal{I}\right|=1$ in terms of where
to converge. In terms of algorithmic implementation, K-SVD employs
SVD which gives the optimal solution without any iterations visible
to users. As a comparison, our implementations of SimCO are built
on gradient descent, which is well-known for its potentially slow
convergence rate. Nevertheless, our numerical tests show similar convergence
rates (similar number of iterations) for primitive SimCO and K-SVD.
This implies that the flexibility of updating codewords simultaneously
significantly reduces the number of iterations. 

When $\left|\mathcal{I}\right|=1$, the rank-one matrix approximation
problem arises in both primitive SimCO and K-SVD. Formally, let $\bm{A}\in\mathbb{R}^{m\times n}$
be a matrix, where $m\ge1$ and $n\ge1$ are arbitrary positive integers.
Without loss of generality, assume that $m\le n$. Suppose that the
sorted singular values satisfy $\lambda_{1}>\lambda_{2}\ge\lambda_{3}\ge\cdots\ge\lambda_{m}$.
Define 
\begin{equation}
f\left(\bm{u}\right)=\underset{\bm{w}\in\mathbb{R}^{n}}{\min}\;\left\Vert \bm{A}-\bm{u}\bm{w}^{T}\right\Vert _{F}^{2},\;\forall\bm{u}\in\mathcal{U}_{m,1}.\label{eq:f-u}
\end{equation}
 The rank-one matrix approximation problem can be written as the following
optimization problem 
\begin{equation}
\underset{\bm{u}\in\mathcal{U}_{m,1}}{\min}\; f\left(\bm{u}\right).\label{eq:opt-Stiefel}
\end{equation}

We shall analyze the performance of gradient descent in the rank-one
matrix approximation problem. To avoid numerical problems that may
arise in practical implementations, we consider an ideal gradient
descent procedure with infinitesimal step sizes. (Note that true gradient
descent requires infinitesimal steps.) More specifically, let $\epsilon$
be a positive number. From a given starting point, one takes steps
of size $\epsilon$ along the negative gradient direction until the
objective function stops decreasing. Letting $\epsilon$ approach
zero gives the ideal gradient descent procedure under consideration. 

The following theorem establishes that the described gradient descent
procedure finds the best rank-one approximation with probability one. 
\begin{thm}
\label{thm:Gradient-Works}Consider a matrix $\bm{A}\in\mathbb{R}^{m\times n}$
and its singular value decomposition. Employ the gradient descent
procedure with infinitesimal steps to solve (\ref{eq:f-u}). Suppose
the starting point, denoted by $\bm{u}_{0}$, is randomly generated
from the uniform distribution on $\mathcal{U}_{m,1}$. Then the gradient
descent procedure finds a global minimizer with probability one. 
\end{thm}
\vspace{0.01cm}

The proof is detailed in Appendix  \ref{sub:Find global optimum}. 
\begin{rem}
The notion of Grassmann manifold is essential in the proof. The reason
is that the global minimizer is not unique: if $\bm{u}\in\mathcal{U}_{m,1}$
is a global minimizer, then so is $-\bm{u}$. In other words, only
the subspace spanned by a global minimizer is unique. 
\end{rem}
\vspace{0.00cm}
\begin{rem}
According to the authors' knowledge, this is the first result showing
that a gradient search on Grassmann manifold solves the rank-one matrix
approximation problem. In literature, it has been shown that there
are multiple stationary points for rank-one matrix approximation problem
\cite[Proposition 4.6.2]{Absil2008:Book:OptimizatoinManifolds}. Our
results show that a gradient descent method will not converge to stationary
points other than global minimizers. More recently, the rank-one decomposition
problem where $\lambda_{2}=\lambda_{3}=\cdots=\lambda_{m}=0$ was
studied in \cite{Dai2011:LRMCGeometric}. Our proof technique is significantly
different as the effects of the eigen-spaces corresponding to $\lambda_{2},\cdots,\lambda_{m}$
need to be considered for the rank-one approximation problem.
\end{rem}

\section{\label{sec:Simulations}Empirical Tests}

In this section, we numerically test the proposed primitive and regularized
SimCO. In the test of SimCO, all codewords are updated simultaneously,
i.e., $\mathcal{I}=\left[d\right]$. In Section \ref{sub:ill-condition-case},
we show that MOD%
\footnote{In the tested MOD, the columns in $\bm{D}$ are normalized after each
dictionary update. This extra step is performed because many sparse
coding algorithms requires normalized dictionary. Furthermore, our
preliminary simulations (not shown in this paper) show that the performance
of dictionary update could seriously deteriorate if the columns are
not normalized. %
}, K-SVD, and primitive SimCO may result in an ill-conditioned dictionary
while regularized SimCO can mitigate this problem. Learning performance
of synthetic and real data is presented in Sections \ref{sub:Synthetic-Data}
and \ref{sub:Real-Data} respectively. Running time comparison of
different algorithms is conducted in Section \ref{sub:Running-time}.
Note that SimCO algorithms are implemented by using simple gradient
descent method. Simulation results suggest that simultaneously updating
codewords significantly speeds up the convergence and the regularization
term substantially improves the learning performance.

\subsection{\label{sub:ill-condition-case}Ill-conditioned Dictionaries}

In this subsection, we handpick a particular example to show that
MOD, K-SVD and primitive SimCO may converge to an ill-conditioned
dictionary. In the example, the training samples $\bm{Y}\in\mathbb{R}^{16\times78}$
are computed via $\bm{Y}=\bm{D}_{{\rm true}}\bm{X}_{{\rm true}}$,
where $\bm{D}_{{\rm true}}\in\mathbb{R}^{16\times32}$ , $\bm{X}_{{\rm true}}\in\mathbb{R}^{32\times78}$,
and each column of $\bm{X}$ contains exactly $4$ nonzero components.
We assume that the sparse coding stage is perfect, i.e., $\Omega_{{\rm true}}$
is available. We start with a particular choice of the initial dictionary
$\bm{D}_{0}\in\mathcal{D}$. The regularization constant $\mu$ in
regularized SimCO is set to $\mu=0.01$.

\begin{figure*}
\includegraphics[scale=0.4]{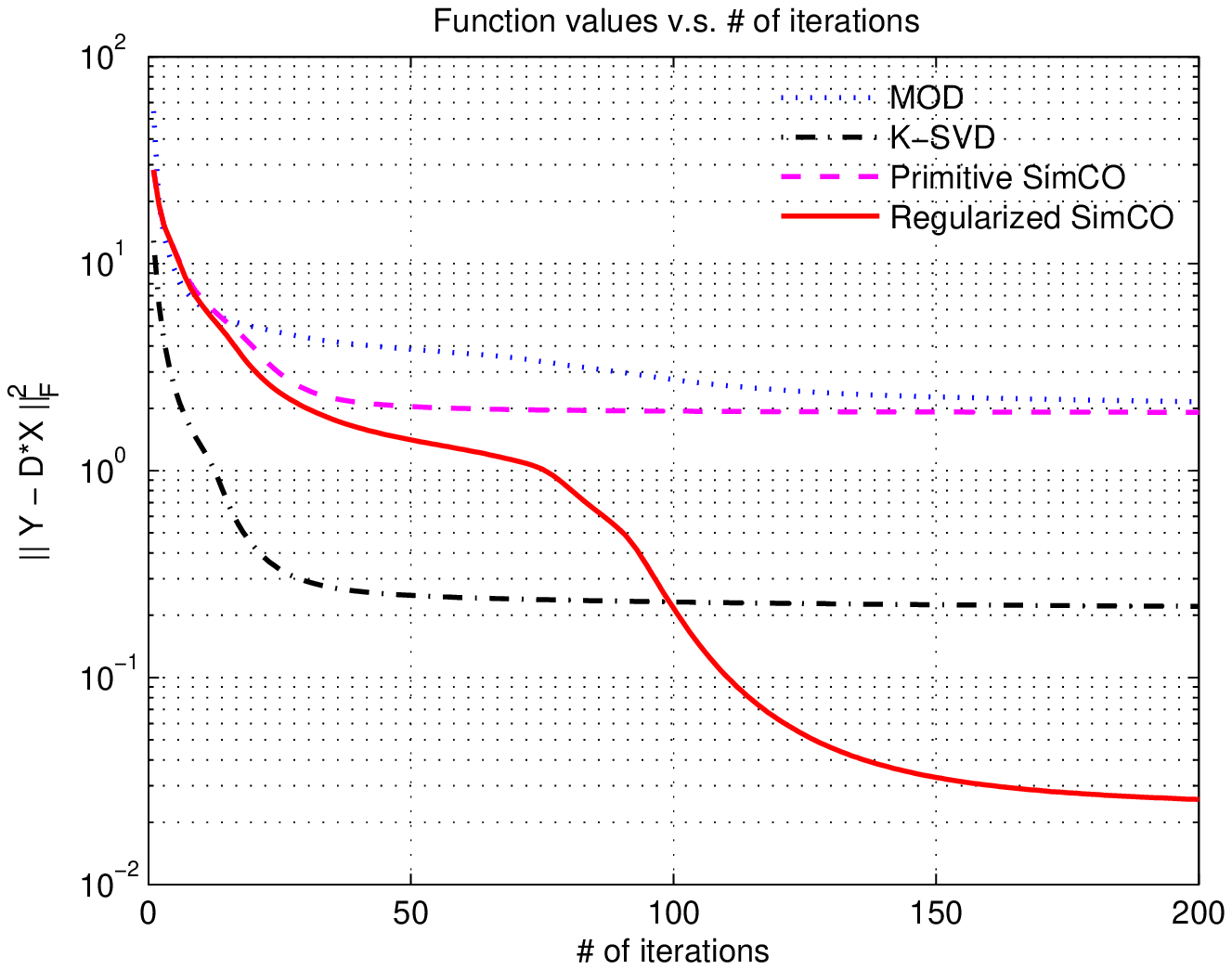}\includegraphics[scale=0.4]{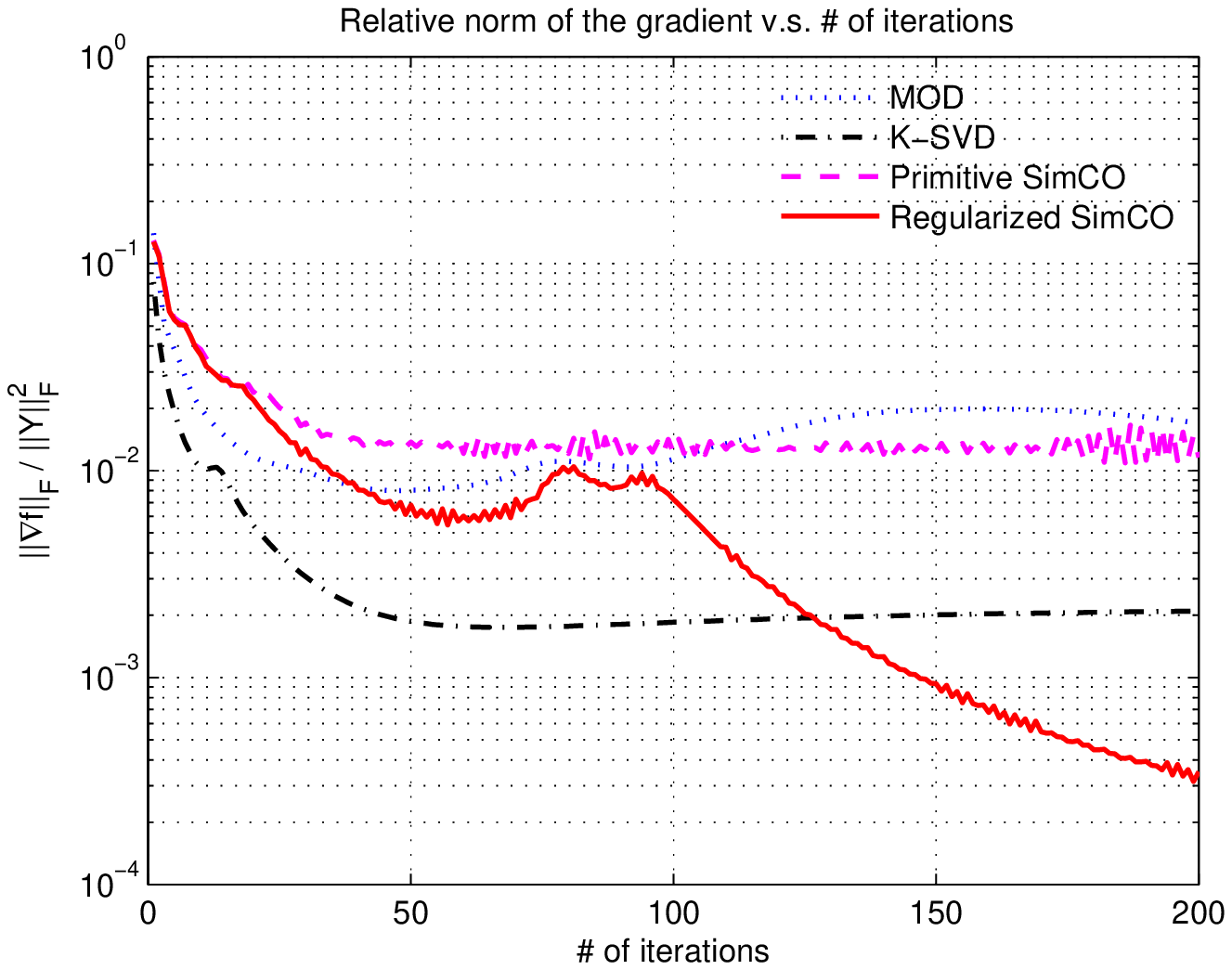}\includegraphics[scale=0.4]{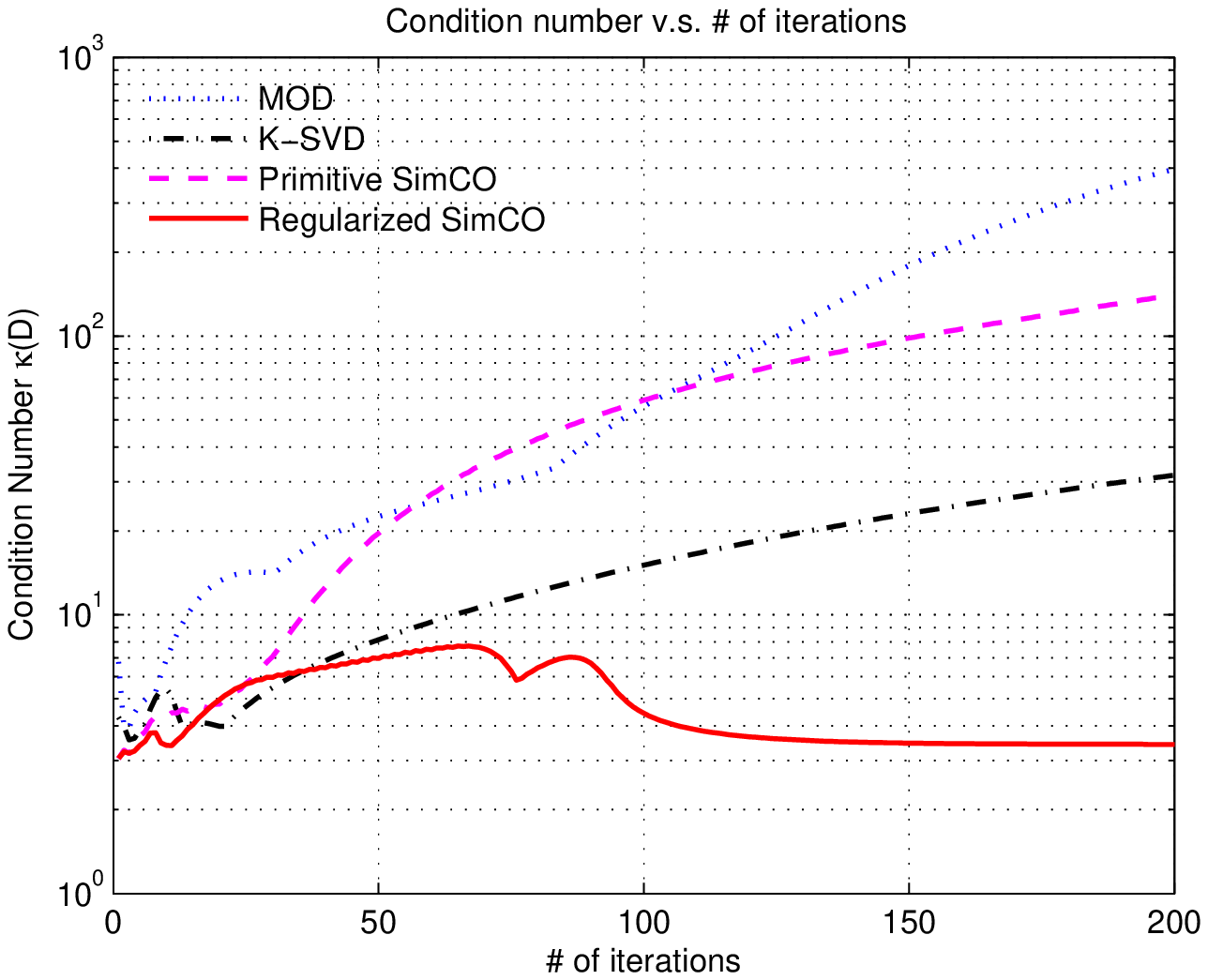}

\caption{\label{fig:ill-condition}Starting with the same point, the convergence
behaviors of MOD, K-SVD, primitive SimCO and regularized SimCO are
different. In this particular example, only regularized SimCO avoids
converging to a singular point. }
\end{figure*}

The numerical results are presented in Figure \ref{fig:ill-condition}.
In the left sub-figure, we compare the learning performance in terms
of $\left\Vert \bm{Y}-\bm{D}\bm{X}\right\Vert _{F}^{2}$. In the middle
sub-figure, we study the behavior of the gradient $\nabla_{\bm{D}}f\left(\bm{D}\right)$
for different algorithms. In the right sub-figure, we depict the condition
number of the dictionary defined as 
\[
\kappa\left(\bm{D}\right)=\underset{1\le j\le d}{\max}\;\lambda_{\max}\left(\bm{D}_{:,\Omega\left(:,j\right)}\right)/\lambda_{\min}\left(\bm{D}_{:,\Omega\left(:,j\right)}\right).
\]
 Here, note that $\kappa\left(\bm{D}_{{\rm true}}\right)=3.39$. The
results in Figure \ref{fig:ill-condition} show that 
\begin{enumerate}
\item When the number of iterations exceeds 50, MOD, K-SVD and primitive
SimCO stop improving the training performance: the value of $f$ decreases
very slowly with further iterations. Surprisingly, the gradients in
these methods do not converge to zero. This implies that these methods
\emph{do not converge to local minimizers}. A more careful study reveals
that these algorithms converge to points where the curvature (Hessian)
of the objective function $f\left(\bm{D}\right)$ is large: the gradient
of the objective function $\nabla_{\bm{D}}f$ changes dramatically
in a small neighborhood. 
\item The above phenomenon can be well explained by checking the ill-condition
of the dictionary. After 100 iterations, the condition number $\kappa\left(\bm{D}\right)$
remains large ($>10$) for MOD, K-SVD, and primitive SimCO. 
\item By adding a regularized term and choosing the regularization constant
properly, regularized SimCO avoids the convergence to an ill-conditioned
dictionary. 
\end{enumerate}
In fact, our simulations in Section \ref{sub:Synthetic-Data} show
that the performance of primitive SimCO is not as good as other methods.
We tracked all the simulated samples and found that it is because
primitive SimCO may converge to a singular point very fast. Adding
the regularization term significantly improves the performance (see
Sections \ref{sub:Synthetic-Data} and \ref{sub:Real-Data}). The
necessity of regularized SimCO is therefore clear.

\subsection{\label{sub:Synthetic-Data}Experiments on Synthetic Data}

The setting for synthetic data tests is summarized as follows. The
training samples are generated via $\bm{Y}=\bm{D}_{{\rm true}}\bm{X}_{{\rm true}}$.
Here, the columns of $\bm{D}_{{\rm true}}$ are randomly generated
from the uniform distribution on the Stiefel manifold $\mathcal{U}_{m,1}$.
Each column of $\bm{X}_{{\rm true}}$ contains exactly $S$ many non-zeros:
the position of the non-zeros are uniformly distributed on the set
${\left[d\right] \choose S}=\left\{ \left\{ i_{1},\cdots,i_{S}\right\} :\;1\le i_{k}\ne i_{\ell}\le d\right\} $;
and the values of the non-zeros are standard Gaussian distributed.
In the tests, we fix $m=16$, $d=32$, and $S=4$, and change $n$,
i.e., the number of training samples. Note that we intentionally choose
$n$ to be small, which corresponds to the challenging case. 

\begin{figure}
\begin{centering}
\subfloat[Noiseless case.]{\begin{centering}
\includegraphics[scale=0.45]{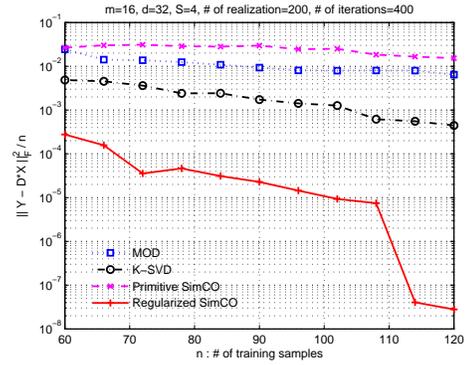}
\par\end{centering}

}
\par\end{centering}

\centering{}\subfloat[Noisy case: SNR of training samples is 20dB. Note that there always
exists a floor in reconstruction error that is proportional to noise.]{\begin{centering}
\includegraphics[scale=0.45]{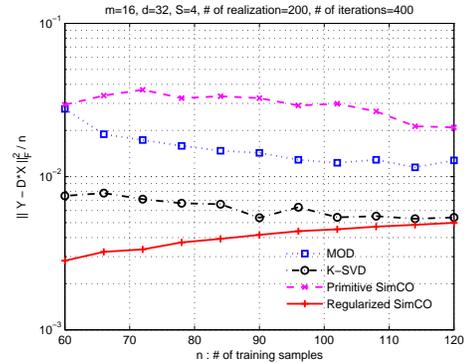}
\par\end{centering}

} \caption{\label{fig:Comparison-No-OMP}Performance comparison of dictionary
update (no sparse coding step).}
\end{figure}

We first focus on the performance of dictionary update by assuming
the true sparsity $\Omega_{{\rm true}}$ is available. Results are
presented in Fig. \ref{fig:Comparison-No-OMP}. Note that the objective
function of regularized SimCO is different from that of other methods.
The ideal way to test regularized SimCO is to sequentially decrease
the regularization constant $\mu$ to zero. In practice, we use the
following simple strategy: the total number of iterations is set to
400; we change $\mu$ from $1e-1$ to $1e-2$, $1e-3$, and $1e-4$,
for every 100 iterations. Simulations show that the average performance
of regularized SimCO is consistently better than that of MOD and K-SVD.
Note that there always exists a floor in reconstruction error that
is proportional to noise. The normalized learning performance $\left\Vert \bm{Y}-\bm{D}\bm{X}\right\Vert _{F}^{2}/n$
is presented in Figure \ref{fig:Comparison-No-OMP}. The average performance
of regularized SimCO is consistently better than that of MOD and K-SVD. 

Then we evaluate the overall dictionary learning performance by combining
the dictionary update and sparse coding stages. For sparse coding,
we adopt the OMP algorithm \cite{Tropp2007} as it has been intensively
used for testing the K-SVD method in \cite{AharonEB2006,Elad2006:ImageDenoisingDictionary}.
The overall dictionary learning procedure is given in Algorithm \ref{alg:dictionary-learning}.
We refer to the iterations between sparse coding and dictionary learning
stages as outer-iterations, and the iterations within the dictionary
update stage as inner-iterations. In our test, the number of outer-iterations
is set to 50, and the number of inner-iterations of is set to 1. Furthermore,
in regularized SimCO, the regularized constant is set to $\mu=1e-1$
during the first 30 outer-iterations, and $\mu=0$ during the rest
20 outer-iterations. The normalized learning performance $\left\Vert \bm{Y}-\bm{D}\bm{X}\right\Vert _{F}^{2}/n$
is depicted in Figure \ref{fig:Comparison-No-OMP}. Again, the average
performance of regularized SimCO is consistently better than that
of other methods. 

\begin{figure}
\centering{}\includegraphics[scale=0.45]{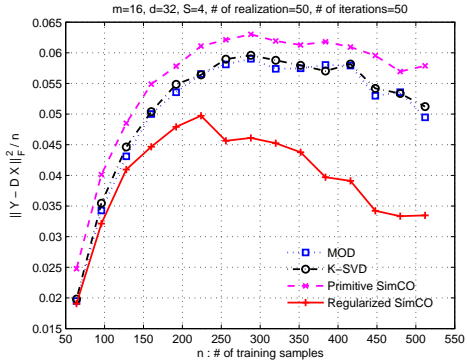} \caption{\label{fig:Comparison-OMP}Performance comparison of dictionary learning
using OMP for sparse coding.}
\end{figure}

Note that in the tests presented in this subsection, the performance
of primitive SimCO is not as good as other methods. This motivates
and justifies regularized SimCO.

\subsection{\label{sub:Real-Data}Numerical Results for Image Denoising}

\begin{figure*}
\raggedright{}\hfill{}\includegraphics[scale=0.45]{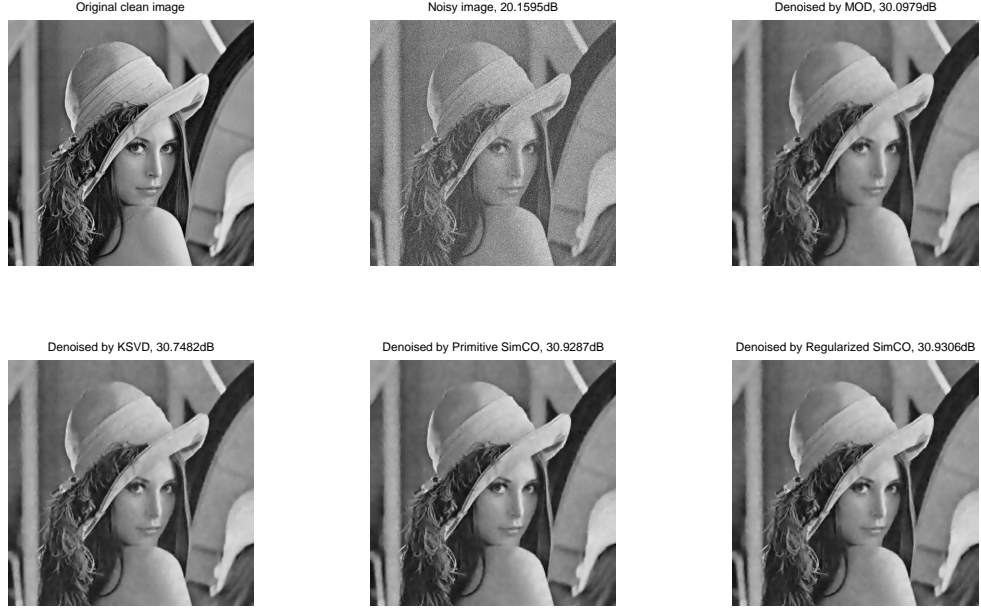}\hfill{}
\caption{\label{fig:RealDataDemo}Example of the image denoising using dictionary
learning. PSNR values in dB are given in sub-figure titles. }
\end{figure*}

As we mentioned in the introduction part, dictionary learning methods
have many applications. In this subsection, we look at one particular
application, i.e., image denoising. Here, a corrupted image with noise
was used to train the dictionary: we take 1,000 (significantly less
than 65,000 used in \cite{Elad2006:ImageDenoisingDictionary}) blocks
(of size $8\times8$) of the corrupted image as training samples.
The number of codewords in the training dictionary is 256. For dictionary
learning, we iterate the sparse coding and dictionary update stages
for 10 times. The sparse coding stage is based on the OMP algorithm
implemented in \cite{Elad2006:ImageDenoisingDictionary}. In the dictionary
update stage, different algorithms are tested. For regularized SimCO,
the regularization constant is set to $\mu=0.05$. During each dictionary
update stage, the line search procedure is only performed once. After
the whole process of dictionary learning, we use the learned dictionary
to reconstruct the image. The reconstruction results are presented
in Fig. \ref{fig:RealDataDemo}. While all dictionary learning methods
significantly improves the image SNRs, the largest gain was obtained
from regularized SimCO. 

\subsection{\label{sub:Running-time}Comments on the Running Time}

We compare the running time of different dictionary update algorithms
in Table \ref{tab:Running-Time}. It is empirically observed that
SimCO runs faster than K-SVD but slower than MOD. The speed-up compared
with K-SVD comes from the simultaneous update of codewords. That SimCO
is slower than MOD is not surprising for the following reasons: MOD
also updates all the codewords simultaneously; and MOD only requires
solving least-squares problems, which are much simpler than the optimization
problem in SimCO. 

\begin{table}

\caption{\label{tab:Running-Time}Comparison of running time (in seconds) for
dictionary learning. Note that sparse coding step is included in producing
Fig. \ref{fig:Comparison-OMP} and \ref{fig:RealDataDemo}.}

\begin{tabular}{|c|>{\centering}m{1.4cm}|>{\centering}m{1.4cm}|>{\centering}m{1.4cm}|>{\centering}m{1.4cm}|}
\hline 
 & MOD & K-SVD & Primitive SimCO & Regularized SimCO\tabularnewline
\hline 
\hline 
Fig. \ref{fig:Comparison-No-OMP}(a) & $2.4\times10^{4}$ & $2.0\times10^{5}$ & $5.1\times10^{4}$ & $4.0\times10^{4}$\tabularnewline
\hline 
Fig. \ref{fig:Comparison-No-OMP}(b) & $2.3\times10^{4}$ & $1.9\times10^{5}$ & $5.0\times10^{4}$ & $4.0\times10^{4}$\tabularnewline
\hline 
Fig. \ref{fig:Comparison-OMP} & $1.5\times10^{4}$ & $3.7\times10^{4}$ & $3.1\times10^{4}$ & $3.1\times10^{4}$\tabularnewline
\hline 
Fig. \ref{fig:RealDataDemo} & 1.42 & 29.43 & 2.63 & 2.72\tabularnewline
\hline 
\end{tabular}
\end{table}

\section{\label{sec:Conclusions}Conclusions}

We have presented a new framework for dictionary update. It is based
on optimization on manifolds and allows a simultaneous update of all
codewords and the corresponding coefficients. Two algorithms, primitive
and regularized SimCO have been developed. On the theoretical aspect,
we have established the equivalence between primitive SimCO and K-SVD
when only one codeword update is considered. On the more practical
side, numerical results are presented to show the good learning performance
and fast running speed of regularized SimCO. 

\appendix

\subsection{\label{sub:Find global optimum}Proof of Theorem \ref{thm:Gradient-Works}}

The following notations are repeatedly used in the proofs. Consider
the singular value decomposition $\bm{A}=\sum_{i=1}^{m}\lambda_{i}\bm{u}_{\bm{A},i}\bm{v}_{\bm{A},i}^{T}$,
where $\lambda_{1}>\lambda_{2}\ge\cdots\ge\lambda_{m}\ge0$ are the
singular values, and $\bm{u}_{\bm{A},i}$ and $\bm{v}_{\bm{A},i}$
are the left and right singular vectors corresponding to $\lambda_{i}$
respectively. It is clear that the objective function $f\left(\bm{u}\right)=\inf_{\bm{w}\in\mathbb{R}^{n}}\;\left\Vert \bm{A}-\bm{u}\bm{w}^{T}\right\Vert _{F}^{2}$
has two global minimizers $\pm\bm{u}_{\bm{A},1}$. For a given $\bm{u}\in\mathcal{U}_{m,1}$,
the angle between $\bm{u}$ and the closest global minimizer is defined
as 
\[
\theta=\mbox{cos}^{-1}\left|\left\langle \bm{u},\bm{u}_{\bm{A},1}\right\rangle \right|.
\]

The crux of the proof is that along the gradient descent path, the
angle $\theta$ is monotonically decreasing. Suppose that the starting
angle is less than $\pi/2$. Then the only stationary points are when
the angle $\theta$ is zero. Hence, the gradient descent search converges
to a global minimizer. The probability one part comes from that the
starting angle equals to $\pi/2$ with probability zero. 

To formalize the idea, it is assumed that the starting point $\bm{u}_{0}\in\mathcal{U}_{m,1}$
is randomly generated from the uniform distribution on the Stiefel
manifold. Define a set $\mathcal{B}\subset\mathcal{U}_{m,1}$ to describe
the set of {}``bad'' starting points. It is defined by 
\[
\mathcal{B}=\left\{ \bm{u}\in\mathcal{U}_{m,1}:\;\bm{u}^{T}\bm{u}_{\bm{A},1}=0\right\} ,
\]
which contains all unit vectors that are orthogonal to $\bm{u}_{\bm{A},1}$.
According to \cite{Dai_IT2008_Quantization_Grassmannian_manifold},
under the uniform measure on $\mathcal{U}_{m,1}$, the measure of
the set $\mathcal{B}$ is zero. As a result, the starting point $\bm{u}_{0}\notin\mathcal{B}$
with probability one. The reason that we refer to $\mathcal{B}$ as
the set of {}``bad'' starting points is explained by the following
lemma.
\begin{lem}
\label{lem:BadSet1}Starting from any $\bm{u}_{0}\in\mathcal{B}$,
a gradient descent path stays in the set $\mathcal{B}$. \end{lem}
\begin{IEEEproof}
This lemma can be proved by computing the gradient of $f$ at a $\bm{u}\in\mathcal{B}$.
Let $\bm{w}_{\bm{u}}\in\mathbb{R}^{n}$ be the optimal solution of
the least squares problem in $f\left(\bm{u}\right)=\inf_{\bm{w}\in\mathbb{R}^{n}}\;\left\Vert \bm{A}-\bm{u}\bm{w}^{T}\right\Vert _{F}^{2}$.
It can be verified that $\bm{w}_{\bm{u}}=\bm{A}^{T}\bm{u}$ and $\nabla f=-2\left(\bm{A}-\bm{u}\bm{w}_{\bm{u}}^{T}\right)\bm{w}_{\bm{u}}$.
It is clear that 
\begin{align*}
\nabla f & =-2\left(\bm{A}-\bm{u}\bm{w}_{\bm{u}}^{T}\right)\bm{w}_{\bm{u}}=-2\left(\bm{A}-\bm{u}\bm{u}^{T}\bm{A}\right)\bm{A}^{T}\bm{u}\\
 & =-2\sum_{i}\lambda_{i}^{2}\bm{u}_{\bm{A},i}\bm{u}_{\bm{A},i}^{T}\bm{u}+2\bm{u}\left(\bm{u}^{T}\bm{A}\bm{A}^{T}\bm{u}\right)
\end{align*}
 When $\bm{u}_{0}\in\mathcal{B}$, it holds that $\left\langle \bm{u}_{0},\bm{u}_{\bm{A},1}\right\rangle =0$
and $\left\langle \nabla f\left(\bm{u}_{0}\right),\bm{u}_{\bm{A},1}\right\rangle =0$.
Since both $\bm{u}_{0}$ and the gradient descent direction are orthogonal
to $\bm{u}_{\bm{A},1}$, the gradient descent path starting from $\bm{u}_{0}\in\mathcal{B}$
stays in $\mathcal{B}$.  
\end{IEEEproof}
\vspace{0cm}

Now consider a starting points $\bm{u}_{0}\notin\mathcal{B}$. We
shall show that the angle $\theta$ is monotonically decreasing along
the gradient descent path. Towards this end, the notions of directional
derivative play an important role. View $\theta$ as a function of
$\bm{u}\in\mathcal{U}_{m,1}$. The directional derivative of $\theta$
at $\bm{u}\in\mathcal{U}_{m,1}$ along a direction vector $\bm{h}\in\mathbb{R}^{m}$,
denoted by $\nabla_{\bm{h}}\theta\in\mathbb{R}$, is defined as 
\[
\nabla_{\bm{h}}\theta=\underset{\epsilon\rightarrow0}{\lim}\frac{\theta\left(\bm{u}+\epsilon\bm{h}\right)-\theta\left(\bm{u}\right)}{\epsilon}.
\]
Note the relationship between directional derivative and gradient
given by $\nabla_{\bm{h}}\theta=\left\langle \nabla\theta,\bm{h}\right\rangle $.
With this definition, the following lemma plays the central role in
establishing Theorem \ref{thm:Gradient-Works}. 
\begin{lem}
\label{lem:theta-decrease}Consider a $\bm{u}\in\mathcal{U}_{m,1}$
such that $\theta\left(\bm{u}\right):=\cos^{-1}\left(\left|\left\langle \bm{u},\bm{u}_{\bm{A},1}\right\rangle \right|\right)\in\left(0,\pi/2\right)$.
Let $\bm{h}_{f}=-\nabla f\left(\bm{u}\right)$ be the gradient of
the objective function $f$ at $\bm{u}$. Then it holds $\nabla_{\bm{h}_{f}}\theta<0$. 
\end{lem}
\vspace{0cm}The proof of this lemma is detailed in Appendix \ref{sub:Proof-of-theta-decrease}. 

The implications of this lemma are twofold. First, it implies that
$\bm{h}_{f}=-\nabla f\ne\bm{0}$ for all $\bm{u}$ such that $\theta\left(\bm{u}\right)\in\left(0,\pi/2\right)$.
Hence, the only possible stationary points in $\mathcal{U}_{m,1}\backslash\mathcal{B}$
are $\bm{u}_{\bm{A},1}$ and $-\bm{u}_{\bm{A},1}$. Second, starting
from $\bm{u}_{0}\in\mathcal{B}$, the angle $\theta$ decreases along
the gradient descent path. As a result, a gradient descent path will
not enter $\mathcal{B}$. It will converge to $\bm{u}_{\bm{A},1}$
or $-\bm{u}_{\bm{A},1}$. Theorem \ref{thm:Gradient-Works} is therefore
proved.

\subsection{\label{sub:Proof-of-theta-decrease}Proof of Lemma \ref{lem:theta-decrease}}

This appendix is devoted to prove Lemma \ref{lem:theta-decrease},
i.e., $\nabla_{\bm{h}_{f}}\theta<0$. Note that $\nabla_{\bm{h}_{f}}\theta=\left\langle \bm{h}_{f},\nabla\theta\right\rangle =\left\langle -\nabla f,\nabla\theta\right\rangle =\nabla_{-\nabla\theta}f$.
It suffices to show that $\nabla_{-\nabla\theta}f<0$. 

\begin{figure}
\hfill{}\includegraphics[scale=0.35]{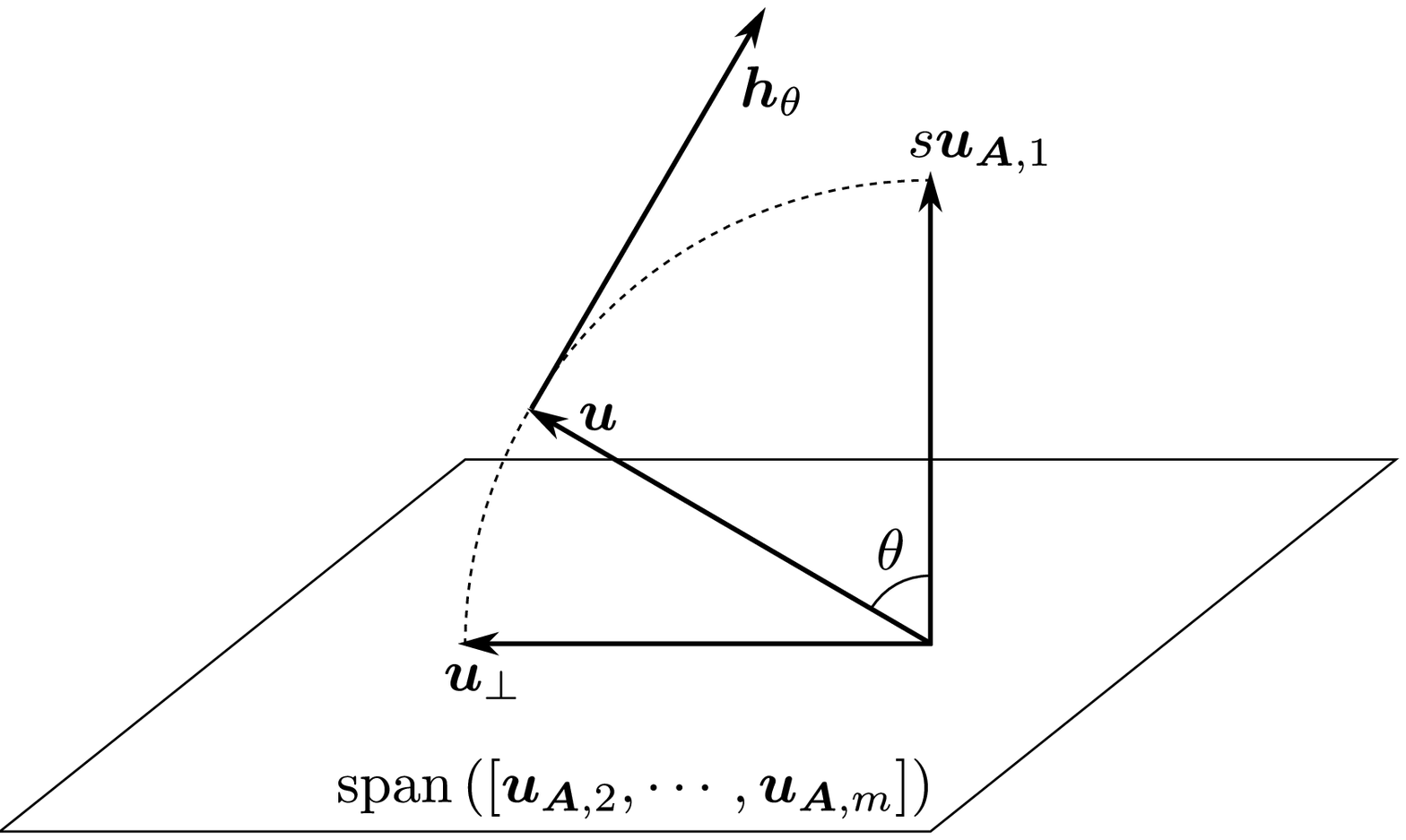}\hfill{}

\caption{\label{fig:Illustration-Vectors}Illustration of $\bm{u}$, $\bm{u}_{\bm{A},1}$,
$\bm{h}_{\theta}$ and $\bm{u}_{\perp}$.}
\end{figure}

Towards this end, the following definitions are useful. Define $s=\mbox{sign}\left(\bm{u}^{T}\bm{u}_{\bm{A},1}\right)$.
Then the vector $s\bm{u}_{\bm{A},1}$ is one of the two global minimizers
that is the closest to $\bm{u}$. It can be also verified that $\theta=\cos^{-1}\left\langle \bm{u},s\bm{u}_{\bm{A},1}\right\rangle $.
Furthermore, suppose that $\theta\in\left(0,\pi/2\right)$. Define
\[
\bm{h}_{\theta}=\frac{s\bm{u}_{\bm{A},1}-\bm{u}\cos\theta}{\sin\theta},\;{\rm and}\;\bm{u}_{\perp}=\frac{\bm{u}-s\bm{u}_{\bm{A},1}\cos\theta}{\sin\theta}.
\]
Clearly, vectors $\bm{h}_{\theta}$ and $\bm{u}_{\perp}$ are well-defined
when $\theta\in\left(0,\pi/2\right)$. The relationship among $\bm{u}$,
$\bm{u}_{\bm{A},1}$, $\bm{h}_{\theta}$ and $\bm{u}_{\perp}$ is
illustrated in Figure \ref{fig:Illustration-Vectors}. Intuitively,
the vector $\bm{h}_{\theta}$ is the tangent vector that pushes $\bm{u}$
towards the global minimizer $s\bm{u}_{\bm{A},1}$. 

In the following, we show that $\nabla_{-\nabla\theta}f=\nabla_{\bm{h}_{\theta}}f$
if we restrict $\bm{u}\in\mathcal{U}_{m,1}$. By the definition of
the directional derivative, one has%
\footnote{The denominator comes from the restriction that  $\bm{u}\in\mathcal{U}_{m,1}$.%
} 
\begin{align*}
\nabla_{-\nabla\theta}\bm{u} & =\underset{\epsilon\rightarrow0}{\lim}\frac{\bm{u}-\epsilon\nabla\theta}{\left\Vert \bm{u}-\epsilon\nabla\theta\right\Vert }.
\end{align*}
Note that 
\begin{align*}
\nabla\theta & =\nabla\left(\cos^{-1}\left(\cos\theta\right)\right)\\
 & =-\frac{1}{\sqrt{1-\cos^{2}\theta}}\nabla\left\langle \bm{u},s\bm{u}_{\bm{A},1}\right\rangle =-\frac{1}{\sin\theta}\left(s\bm{u}_{\bm{A},1}\right).
\end{align*}
Since $s\bm{u}_{\bm{A},1}=\bm{u}\cos\theta+\bm{h}_{\theta}\sin\theta$,
one has 
\begin{align*}
\bm{u}-\epsilon\nabla\theta & =\bm{u}+\frac{\epsilon}{\sin\theta}\left(s\bm{u}_{\bm{A},1}\right)\\
 & =\bm{u}\left(1+\epsilon\cos\theta/\sin\theta\right)+\epsilon\bm{h}_{\theta}.
\end{align*}
Substitute it back to $\nabla_{-\nabla\theta}\bm{u}$. One has $\nabla_{-\nabla\theta}\bm{u}=\bm{h}_{\theta}.$
In other words, if $\bm{u}\in\mathcal{U}_{m,1}$, then $\nabla_{-\nabla\theta}f=\nabla_{\bm{h}_{\theta}}f.$

To compute $\nabla_{\bm{h}_{\theta}}f$, note that $f\left(\bm{u}\right)=\left\Vert \bm{A}-\bm{u}\bm{w}_{\bm{u}}^{T}\right\Vert _{F}^{2}=\left\Vert \bm{A}\right\Vert _{F}^{2}-\left\Vert \bm{u}^{T}\bm{A}\right\Vert _{2}^{2}.$
Now define 
\[
g\left(\bm{u}\right)=\left\Vert \bm{u}^{T}\bm{A}\right\Vert _{2}^{2}.
\]
Then clearly $\nabla_{\bm{h}_{\theta}}f=-\nabla_{\bm{h}_{\theta}}g$.
To proceed, we also decompose $\bm{A}$ as follows. Recall the SVD
of $\bm{A}$ given by $\bm{A}=\sum_{i=1}^{m}\lambda_{i}\bm{u}_{\bm{A},i}\bm{v}_{\bm{A},i}^{T}$.
Let $\bm{U}_{\bm{A},\perp}\in\mathcal{U}_{m,m-1}$ contain the left
singular vectors corresponding to $\lambda_{2},\cdots,\lambda_{m}$,
i.e., $\bm{U}_{\bm{A},\perp}=\left[\bm{u}_{\bm{A},2},\cdots,\bm{u}_{\bm{A},m}\right]$.
Similarly define $\bm{V}_{\bm{A},\perp}$. Then, 
\begin{align*}
\bm{A} & =\left[\bm{u}_{\bm{A},1},\bm{U}_{\bm{A},\perp}\right]\mbox{diag}\left(\left[\lambda_{1},\cdots,\lambda_{m}\right]\right)\left[\begin{array}{c}
\bm{v}_{\bm{A},1}^{T}\\
\bm{V}_{\bm{A},\perp}^{T}
\end{array}\right]\\
 & =\left[\bm{u}_{\bm{A},1},\bm{U}_{\bm{A},\perp}\right]\left[\bm{w}_{\bm{A},1},\bm{W}_{\bm{A},\perp}\right]^{T},
\end{align*}
 where $\bm{w}_{\bm{A},i}=\lambda_{i}\bm{v}_{\bm{A},i}$ for $i=1,\cdots,m$,
and $\bm{W}_{\bm{A},\perp}=\left[\bm{w}_{\bm{A},2},\cdots,\bm{w}_{\bm{A},m}\right]$.
It is straightforward to verify that $\bm{w}_{\bm{A}}^{T}\bm{W}_{\bm{A},\perp}=\bm{0}$.

The function $g\left(\bm{u}\right)$ can be decomposed into two parts.
Note that 
\begin{align*}
g\left(\bm{u}\right) & =\left\Vert \bm{u}^{T}\left[\bm{u}_{\bm{A},1},\bm{U}_{\bm{A},\perp}\right]\left[\bm{w}_{\bm{A},1},\bm{W}_{\bm{A},\perp}\right]^{T}\right\Vert _{2}^{2}\\
 & =\left\Vert \bm{u}^{T}\bm{u}_{\bm{A},1}\bm{w}_{\bm{A},1}^{T}\right\Vert _{2}^{2}+\left\Vert \bm{u}^{T}\bm{U}_{\bm{A},\perp}\bm{W}_{\bm{A},\perp}^{T}\right\Vert _{2}^{2}\\
 & \quad+2\left\langle \bm{u}^{T}\bm{u}_{\bm{A},1}\bm{w}_{\bm{A},1}^{T},\bm{u}^{T}\bm{U}_{\bm{A},\perp}\bm{W}_{\bm{A},\perp}^{T}\right\rangle \\
 & =\left\Vert \bm{u}^{T}\bm{u}_{\bm{A},1}\bm{w}_{\bm{A},1}^{T}\right\Vert _{2}^{2}+\left\Vert \bm{u}^{T}\bm{U}_{\bm{A},\perp}\bm{W}_{\bm{A},\perp}^{T}\right\Vert _{2}^{2},
\end{align*}
where the last equality follows from that $\bm{W}_{\bm{A},\perp}^{T}\bm{w}_{\bm{A}}=\bm{0}$
and hence
\[
\left\langle \bm{u}^{T}\bm{u}_{\bm{A},1}\bm{w}_{\bm{A},1}^{T},\bm{u}^{T}\bm{U}_{\bm{A},\perp}\bm{W}_{\bm{A},\perp}^{T}\right\rangle =0.
\]
To further simplify $g\left(\bm{u}\right)$, note that $\cos\theta=\left|\bm{u}^{T}\bm{u}_{\bm{A}}\right|$.
Furthermore, it is straightforward to verify that the projection of
$\bm{u}$ on $\mbox{span}\left(\bm{U}_{\bm{A},\perp}\right)$ is given
by $\bm{U}_{\bm{A},\perp}\bm{U}_{\bm{A},\perp}^{T}\bm{u}=\bm{u}_{\perp}\sin\theta$.
Define $\bm{u}_{R}=\bm{U}_{\bm{A},\perp}^{T}\bm{u}_{\perp}\in\mathbb{R}^{m-1}$.
Then, $\left\Vert \bm{u}_{R}\right\Vert =1$ and 
\begin{align*}
 & \left\Vert \bm{u}^{T}\bm{U}_{\bm{A},\perp}\bm{W}_{\bm{A},\perp}^{T}\right\Vert _{2}^{2}\\
 & =\sin^{2}\theta\left\Vert \bm{u}_{\perp}^{T}\bm{U}_{\bm{A},\perp}\bm{W}_{\bm{A},\perp}^{T}\right\Vert _{2}^{2}\\
 & =\sin^{2}\theta\bm{u}_{R}^{T}\mbox{diag}\left(\left[\lambda_{2}^{2},\cdots,\lambda_{m}^{2}\right]\right)\bm{u}_{R}^{T}.
\end{align*}
Hence, 
\[
g\left(\bm{u}\right)=\cos^{2}\theta\cdot\lambda_{1}+\sin^{2}\theta\bm{u}_{R}^{T}\mbox{diag}\left(\left[\lambda_{2}^{2},\cdots,\lambda_{m}^{2}\right]\right)\bm{u}_{R}.
\]

It is now ready to decide the sign of $\nabla_{\bm{h}_{\theta}}g$.
It is straightforward to verify that 
\[
\nabla_{\bm{h}_{\theta}}\cos\theta=\underset{\epsilon\rightarrow0}{\lim}\left\langle \frac{\bm{u}+\epsilon\bm{h}_{\theta}}{\sqrt{1+\epsilon^{2}}},s\bm{u}_{\bm{A},1}\right\rangle =\sin\theta,
\]
and similarly $\nabla_{\bm{h}_{\theta}}\sin\theta=-\cos\theta.$ Therefore,
\begin{align*}
\nabla_{\bm{h}_{\theta}}\bm{u}_{\perp} & =\nabla_{\bm{h}_{\theta}}\left(\frac{\bm{u}--\cos\theta s\bm{u}_{\bm{A},1}}{\sin\theta}\right)\\
 & =\frac{\bm{h}_{\theta}\sin\theta+\bm{u}\cos\theta-s\bm{u}_{\bm{A},1}}{\sin^{2}\theta}\\
 & =\frac{s\bm{u}_{\bm{A},1}-s\bm{u}_{\bm{A},1}}{\sin^{2}\theta}=\bm{0},
\end{align*}
and $\nabla_{\bm{h}_{\theta}}\bm{u}_{R}=\nabla_{\bm{h}_{\theta}}\left(\bm{U}_{\bm{A},\perp}^{T}\bm{u}_{\perp}\right)=\bm{0}$.
Hence, one has 
\begin{align*}
\nabla_{\bm{h}_{\theta}}g & =\sin2\theta\left(\lambda_{1}-\bm{u}_{R}^{T}\mbox{diag}\left(\left[\lambda_{2}^{2},\cdots,\lambda_{m}^{2}\right]\right)\bm{u}_{R}\right).
\end{align*}
Note that 
\begin{align*}
 & \bm{u}_{R}^{T}\mbox{diag}\left(\left[\lambda_{2}^{2},\cdots,\lambda_{m}^{2}\right]\right)\bm{u}_{R}\\
 & \le\bm{u}_{R}^{T}\mbox{diag}\left(\left[\lambda_{2}^{2},\cdots,\lambda_{2}^{2}\right]\right)\bm{u}_{R}=\lambda_{2}<\lambda_{1}.
\end{align*}
It can be concluded that when $\theta\in\left(0,\pi/2\right)$, $\nabla_{\bm{h}_{\theta}}g>0$
and $\nabla_{\bm{h}_{\theta}}f=-\nabla_{\bm{h}_{\theta}}g<0$. Lemma
\ref{lem:theta-decrease} is therefore proved. 

\bibliographystyle{IEEEtran}
\bibliography{main}

\end{document}